\pdfoutput=1

\documentclass[11pt]{article}

\usepackage{acl}

\usepackage{listings}

\usepackage{times}
\usepackage{latexsym}
\usepackage{algorithm}
\usepackage{hyperref}

\usepackage[T1]{fontenc}

\usepackage[utf8]{inputenc}

\usepackage{microtype}

\usepackage{graphicx}
\usepackage{amssymb}%
\usepackage{pifont}%
\usepackage{amsmath}

\usepackage{caption}
\usepackage{subcaption}

\usepackage{ifthen}
\newboolean{showcomments}
\setboolean{showcomments}{true}

\ifthenelse{\boolean{showcomments}}
{
\newcommand{\florian}[1]{\textcolor{cyan}{[FMai: #1]}}
\newcommand{\todo}[1]{\textcolor{red}{[Todo: #1]}}
\newcommand{\fabio}[1]{\textcolor{orange}{[Fabio: #1]}}
\newcommand{\ff}[1]{\textcolor{purple}{[FF: #1]}}
\newcommand{\james}[1]{\textcolor{magenta}{[JH: #1]}}
\newcommand{\arnaud}[1]{\textcolor{teal}{[Arnaud: #1]}}
\newcommand{\haolin}[1]{\textcolor{blue}{[Haolin: #1]}}
}
{
\newcommand{\florian}[1]{}
\newcommand{\todo}[1]{}
\newcommand{\ff}[1]{}
\newcommand{\james}[1]{}
\newcommand{\fabio}[1]{}
\newcommand{\arnaud}[1]{}
\newcommand{\haolin}[1]{}
}

\newboolean{short}
\setboolean{short}{true}
\newcommand{\short}[2]{\ifthenelse{\boolean{short}}{#1}{#2}}

\newcommand{\hypothesisone}{\textbf{H1}}
\newcommand{\hypothesistwo}{\textbf{H2}}
\newcommand{\hypothesisthree}{\textbf{H3}}

\definecolor{greencolor}{RGB}{35, 107, 65}
\definecolor{grey}{RGB}{47,79,79} %
\definecolor{lightgreen}{RGB}{223, 242, 227}
\definecolor{green}{RGB}{0, 167, 159}
\definecolor{darkgreen}{RGB}{0, 97, 92}
\definecolor{hyperlightgreen}{RGB}{223,242,227}
\definecolor{hypergreen}{RGB}{121,242,146}
\definecolor{hyperdarkgreen}{RGB}{40, 167, 159}
\definecolor{hyperlightorange}{RGB}{223,242,227}
\definecolor{hyperorange}{RGB}{255,130,0}
\definecolor{hyperblue}{RGB}{0,104,252}
\definecolor{hyperyellow}{RGB}{255,190,0}
\newcommand{\green}{Green AI}
\newcommand{\tmmlp}{\operatorname{TM-MLP}}

\title{HyperMixer: An MLP-based Low Cost Alternative to Transformers}

\author{Florian Mai$^{\dagger}$$^{\spadesuit}$  \quad Arnaud Pannatier$^{\dagger}$$^{\spadesuit}$ \quad Fabio Fehr$^{\dagger}$$^{\spadesuit}$ \quad Haolin Chen$^{\dagger}$$^{\spadesuit}$ \\ \bf{François Marelli}$^{\dagger}$$^{\spadesuit}$ \quad \bf{François Fleuret}$^{\clubsuit}$$^{\spadesuit}$$^{\dagger}$ \quad \bf{James Henderson}$^{\dagger}$ \\ 
    $^{\dagger}$Idiap Research Institute, Martigny, Switzerland \\
    $^{\spadesuit}$EPFL, Lausanne, Switzerland \\
    $^{\clubsuit}$University of Geneva, Geneva, Switzerland
    }

\begin{document}
\maketitle

\begin{abstract}
Transformer-based architectures are the model of choice for natural language understanding, but they come at a significant cost, as they have quadratic complexity in the input length, require a lot of training data, and can be difficult to tune.
In the pursuit of lower costs, we investigate simple MLP-based architectures. We find that existing architectures such as MLPMixer, which achieves token mixing through a static MLP applied to each feature independently, are too detached from the inductive biases required for natural language understanding. In this paper, we propose a simple variant, \emph{HyperMixer}, which forms the token mixing MLP dynamically using hypernetworks. 
Empirically, we demonstrate that our model performs better than alternative MLP-based models, and on par with Transformers. In contrast to Transformers, HyperMixer achieves these results at substantially lower costs in terms of processing time, training data, and hyperparameter tuning\footnote{Code is available at \href{https://github.com/idiap/hypermixing}{https://github.com/idiap/hypermixing}}.
\end{abstract}

\section{Introduction}\label{sec:introduction}
Attention-based architectures, such as the Transformer~\citep{vaswani2017attention}, have accelerated the progress in many natural language understanding tasks. Part of their success is a result of a parallelizable training scheme over the input length. This improves training times and allows for larger volumes of data which makes these models amenable to pretraining~\citep{radford2018improving, devlin2018bert}.
Therefore, many current state-of-the-art models are fine-tuned extensions of large pretrained Transformers~\citep{bommasani2021opportunities}.

However, these models come at a significant computational cost. They require considerable resources for pretraining and fine-tuning, which induces high energy consumption~\citep{strubell2019energy} and limits access to research~\citep{bommasani2021opportunities}. Subsequently, \citet{schwartz2020green} argue the need for \emph{"\green"}. They propose a cost evaluation of a result $R$ as following:
\begin{equation*}
Cost (R) \propto E \cdot D \cdot H, \label{eq:cost}
\end{equation*}
where $E$ is the computational cost measured in floating point operations (FPO) of a single example, $D$ is the dataset size, and $H$ is the number of hyperparameter configurations required during tuning.

\begin{figure*}[!h]
  \includegraphics[width=\textwidth]{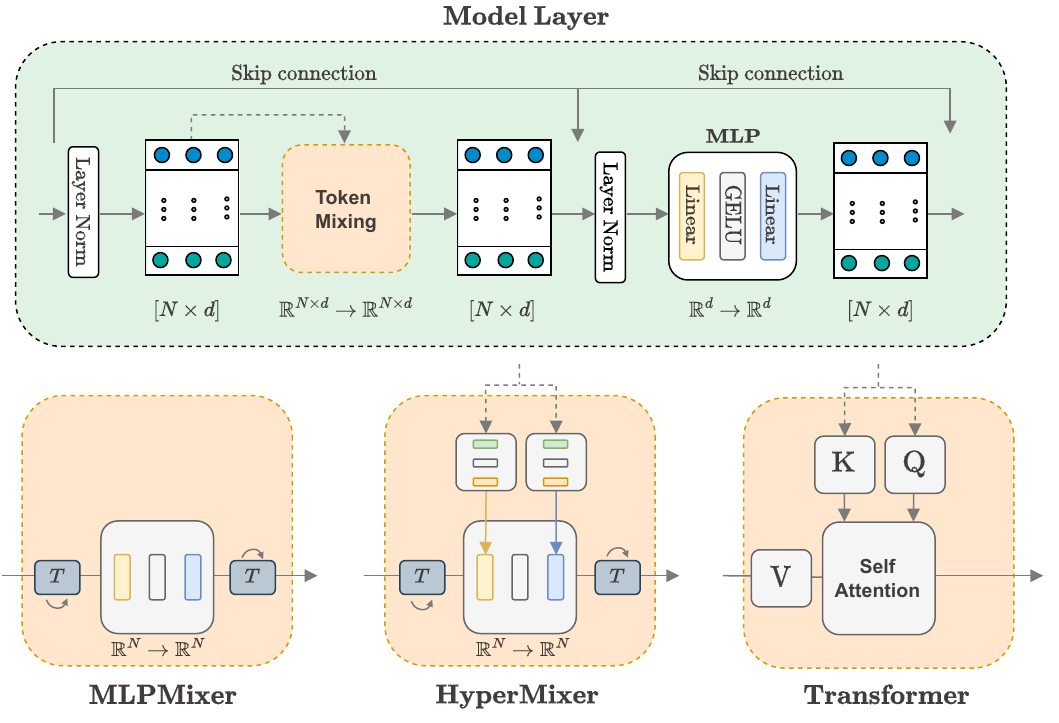}
  \caption{The figure outlines a general model layer consisting of a token mixing component and a feature mixing component (MLP). For token mixing, MLPMixer uses an MLP with a \emph{fixed} size, maximum input length $N$ and \emph{position-specific} weights. In contrast, HyperMixer generates an appropriately sized MLP based on the \emph{variable} size of the input in a \emph{position-invariant} way, similar to the attention mechanism. When using attention as token mixing the whole layer is equivalent to a Transformer encoder layer.}
  \label{fig:model-architecture}
\end{figure*}

To achieve a cost reduction, this paper proposes a simpler alternative to Transformers. We take inspiration from the computer vision community, which has recently seen a surge of research on Multi-Layer Perceptrons (MLPs). Most prominently, MLPMixer~\citep{tolstikhin2021mlp}, which is a simple architecture based on two MLPs: one for token mixing and one for feature mixing. However, the token mixing MLP learns a \emph{fixed-size} set of \emph{position-specific} mappings, arguably making MLPMixer's architecture too detached from the inductive biases needed for natural language understanding, in contrast to Transformers~\citep{henderson2020unstoppable}.

In this paper, we propose a simple variant, \emph{HyperMixer}~(Figure \ref{fig:model-architecture}), which creates a token mixing MLP dynamically using hypernetworks~\citep{ha2016hypernetworks}. This variant is more appropriate, as it learns to generate a \emph{variable-size} set of mappings in a \emph{position-invariant} way, similar to the attention mechanism in Transformers~\citep{vaswani2017attention}. In contrast to Transformer's quadratic complexity, HyperMixer's complexity is linear in the input length. This makes it a competitive alternative for training on longer inputs.

Empirically, we demonstrate that HyperMixer works substantially better on natural language understanding tasks than the original MLPMixer and related alternatives.
In comparison to Transformers, HyperMixer achieves competitive or improved results at a substantially lower cost $Cost (R) \propto E \cdot D \cdot H$: improved inference speeds (E), especially for long inputs; favorable performance in the low-resource regime (D); and efficient tuning for hyperparameters (H).
We attribute HyperMixer's success to its ability to approximate an attention-like function. Further experiments on a synthetic task demonstrate that HyperMixer indeed learns to attend to tokens in similar pattern to the attention mechanism.

In summary, our contributions can be enumerated as follows:
\begin{enumerate}
    \item A novel all-MLP model, HyperMixer, with inductive biases similar to Transformers. (Section: \ref{sec:method})
    \item A performance analysis of HyperMixer against alternative token mixing methods based on controlled experiments on the GLUE benchmark. (Section: \ref{sec:results:peakperformance})  
    \item A comprehensive comparison of the cost $Cost(R)$ of HyperMixer and Transformers. (Sections: \ref{sec:results:time-per-example},~\ref{sec:results:low-resource-performance},~\ref{sec:results:ease-of-tuning})
    \item An ablation demonstrating that HyperMixer learns attention patterns similar to Transformers. (Section: \ref{sec:results:locmix-attention})
\end{enumerate}

\section{Method}\label{sec:method}

\subsection{Inductive Biases in NLP Models}
In machine learning, the inductive biases of a model reflect implicit modeling assumptions which are key to facilitate learning and improve generalization on specific tasks.
In NLP, well-known models with strong inductive biases include: recurrent neural networks~\citep{elman1990finding}, which assume the input to be a sequence; and recursive neural networks~\citep{socher2013recursive}, which assume a tree-structure. While both these inductive biases are reasonable, empirically, Transformers have been more successful in recent years. Furthermore, we reiterate the arguments of \citet{henderson2020unstoppable} for inductive biases in language and apply them to our model design.
\citet{henderson2020unstoppable} attributes the Transformer's success to two concepts: \emph{variable binding} and \emph{systematicity}. 
Variable binding refers to the model's ability to represent multiple entities at once. This is arguably challenging in single-vector representations such as recurrent neural networks. However, Transformers represent each token with its own vector which accounts for variable binding as each token can be interpreted as an entity. Systematicity refers to the models ability to learn generalizable rules that reflect the structural relationship between entities~\citep{fodor1988connectionism}.
Transformers achieve systematicity through the attention mechanism which is a learnable set of functions that determines the interaction between entities by matching query representations to key representations (as shown in Figure~\ref{fig:model-architecture}). The mechanism \emph{modulates}, for every position in the sequence, how to functionally process any other position. Moreover, these function parameters are learnable and shared across all entities.

\subsection{MLPMixer}\label{sec:methods:mlpmixer}
A general layer of MLPMixer is shown in Figure~\ref{fig:model-architecture}. Similarly to Transformers, each token is represented as a vector of features, which undergo (non-linear) transformations in multiple layers. MLPMixer employs two MLPs at each layer, one for \emph{feature mixing} and one for \emph{token mixing}. The feature mixing component is applied to each token vector independently, which models the interactions between features. The Token Mixing MLP ($\tmmlp$) is applied to each feature independently (i.e.\ its vector of values across tokens), which models the interactions between spatial locations or positions. This could be interpreted as a global attention mechanism which is static and position-modulated. Practically, this is achieved by transposing the dimension representing the features and the dimension representing the positions. 
Each vector $\textbf{x}^T_{i} \in \mathbb{R}^{N}$, representing feature $i \leq d$, of some input of fixed length $N$, is input into $\tmmlp$, which has the following form:
\begin{equation}
    \tmmlp(\textbf{x}^T_{i}) = \textbf{W}_1 (\sigma (\textbf{W}_2^{T} \textbf{x}^T_{i})),\label{eq:mlp1}
\end{equation}
where $\textbf{W}_1, \textbf{W}_2 \in \mathbb{R}^{N \times d'}$, and $\sigma$ represents the  $\operatorname{GELU}$ non-linearity~\citep{hendrycks2016gaussian}.
Finally, to facilitate learning, layer normalization~\citep{ba2016layer} and skip connections~\citep{he2016deep} are added around each MLP, respectively.
How to best arrange these components is still an open question~\citep{wang2019learning, bachlechner2021rezero}. We experiment with different variants in Appendix~\ref{appendix:transformer-layout}. 

\paragraph{Considerations for NLP}
The token mixing MLP assumes an input of fixed dimension, which is necessary as the parameters need to be shared across all examples.
However, unlike images, textual input is generally of a variable dimension. Therefore, to apply MLPMixer to texts of variable length, a simplistic approach is to assume a maximum length (e.g. the maximum in the dataset). Thereafter, all inputs are padded to the maximum length and masks are applied in the token mixing MLP. This model is able to do variable binding, since each token is represented by its own vector. However, this model lacks systematicity because the rules learned to model interactions between tokens (i.e. the MLP's weights) are not shared across positions.

\subsection{HyperMixer}

\begin{algorithm}
\begin{lstlisting}[language=Python,  
    commentstyle=\color{blue},
    keywordstyle=\color{hyperorange},  
basicstyle=\ttfamily\tiny\color{black}\bfseries,framesep=10pt,
    escapechar=@]
class HyperMixing(nn.Module):
    def __init__(self, d, @d'@):

        # learnable parameters
        self.hypernetwork_in = MLP([d, d, @d'@])
        self.hypernetwork_out = MLP([d, d, @d'@])

        # layer normalization improves training stability
        self.layer_norm = LayerNorm(d)

    def forward(self, queries, keys, values):
        # queries: [B, M, d]
        # keys / values: [B, N, d]

        # add token information (e.g. position embeddings)
        hyp_in = add_token_information(keys)
        hyp_out = add_token_information(queries)
        
        W1 = self.hypernetwork_in(hyp_in) # [B, N, d@'@]
        W2 = self.hypernetwork_out(hyp_out) # [B, M, d@'@]

        # TM-MLP(x) = W_2 ( GELU ( W_1^T x) )
        # maps [B, d, N] -> [B, d, M]
        token_mixing_mlp = compose_TM_MLP(W1, W2)

        # transpose so MLP is applied to sequence dimension
        values = values.transpose(1, 2) # [B, d, N]
    
        output = token_mixing_mlp(values) # [B, d, M]

        # transpose back
        output = output.transpose(1,2) # [B, M, d]

        # optionally apply LayerNorm
        return self.layer_norm(output)
    
\end{lstlisting}
\caption{HyperMixer pseudo-code}
\label{alg:hypermixer}
\end{algorithm}

HyperMixer includes systematicity into the MLPMixer architecture by introducing a novel token mixing mechanism, \emph{HyperMixing}\footnote{HyperMixing is to HyperMixer what self-attention is to Transformer encoders.}, which can be regarded as a drop-in replacement for attention.
For ease of understanding, we provide pseudo-code in Algorithm~\ref{alg:hypermixer}. While the queries, keys, and values in HyperMixing need not be the same, we will assume they are identical in the following formulation.
HyperMixing relies on the use of hypernetworks, which are used to generate the weights $\textbf{W}_1, \textbf{W}_2$ of $\tmmlp$ (Equation~\ref{eq:mlp1}) dynamically as a function of the input. Let $\textbf{x}_{j} \in \mathbb{R}^{d}$, $j \leq N$, where $N$ is the (variable) dimension of the input, represent token $j$ (i.e., query, key, and value). 
\short{$\textbf{W}_1$ and $\textbf{W}_2$ are generated by parameterized functions $h_1, h_2 : \mathbb{R}^{N \times d} \rightarrow \mathbb{R}^{N \times d'}$.}{We use the following parameterized functions:
\begin{equation*}
    \begin{aligned}
        h_1, h_2 &: \mathbb{R}^{N \times d} \rightarrow \mathbb{R}^{N \times d'},
    \end{aligned}
\end{equation*}
to generate $\textbf{W}_1$ and $\textbf{W}_2$, respectively.
}
Theoretically, $h_1$ and $h_2$ could be any function, including sophisticated networks that consider non-linear interactions between tokens, such as the attention mechanism.
However, this would defeat the purpose of our model, which is simplicity.
Therefore, we choose to generate the rows of the weight matrices from each token independently via another MLP. Concretely, a hypernetwork function can be defined as
\begin{equation*}
    h_i(\textbf{x}) = \left(
\begin{array}{c}
\operatorname{MLP^{\textbf{W}_i}}(\textbf{x}_{1} + \textbf{p}_{1}) \\
\vdots \\
\operatorname{MLP^{\textbf{W}_i}}(\textbf{x}_{N} + \textbf{p}_{N})
\end{array}
\right) \in \mathbb{R}^{N \times d'},
\end{equation*}
where $\operatorname{MLP^{\textbf{W}_1}}, \operatorname{MLP^{\textbf{W}_2}} : \mathbb{R}^{d} \rightarrow \mathbb{R}^{d'}$ are themselves multi-layer perceptrons with GELU non-linearity.
$\textbf{p}_j \in \mathbb{R}^{d}$ is a vector that can encode additional information such as the position via absolute position embeddings~\citep{vaswani2017attention}.

Intuitively, for each token $\textbf{x}_{j}$, $h_1$ decides which information to send to the hidden layer of $\tmmlp$, where the information from all tokens are mixed, and $h_2$ decides for each token how to extract information from the hidden layer. Note that, even though $h_1$ and $h_2$ only consider one token at once, non-linear interactions between tokens are still modeled through the hidden layer of $\tmmlp$.

Finally, layer normalization~\citep{ba2016layer} can be applied to the output of $\tmmlp$. We found this helpful to facilitate training with a wide variety of Transformer layouts (Appendix~\ref{appendix:transformer-layout}).

\paragraph{Tying $h_1$ and $h_2$}
In order to reduce the number of parameters and operations in the model, and thereby the complexity, we found it useful to tie $h_1$ and $h_2$ by setting $\textbf{W}_2 = \textbf{W}_1$. %

\paragraph{Considerations for NLP}
In comparison to the MLPMixer defined in Section \ref{sec:methods:mlpmixer}, the use of hypernetworks overcomes two challenges. Firstly, the input no longer has to be of fixed dimensionality. The hypernetwork generates a token mixing MLP of appropriate dimension as a function of the input.
Secondly, the hypernetwork models the interaction between tokens with shared weights across all positions in the input. Hence, systematicity is ensured.

\section{Related Work}

\short{
Research on all-MLP models like MLPMixer~\citep{tolstikhin2021mlp} is widespread in the computer vision community \citep[among many others]{tu2022maxim, yu2021s2mlp, wang2022dynamixeravision}. 
However, they lack some desirable inductive biases for NLP, which we discuss in length in Appendix~\ref{sec:rw-allmlp}.
Specifically, in contrast to HyperMixer, none of the previously proposed methods simultaneously provide \textbf{i}) \emph{position invariance}, which is important for generalization, \textbf{ii}) \emph{adaptive size} for variable-length inputs, \textbf{iii}) a \emph{global receptive field}, which allows interactions to not be limited to small token neighborhoods, \textbf{iv}) \emph{learnabilty} allowing for universal applicablility to various tasks, and \textbf{v}) \emph{dynamicity}, which means that token mixing is a function of the input.
Consequently, only a few works have used MLP-based models as their backbone in NLP tasks. gMLP~\citep{liu2021payattentionto} serves as one of our baselines and pnlp-mixer \cite{fusco2022pnlp} employs standard MLPMixer on top of a novel token embedding method.

Apart from all-MLP models, there is an abundance of research on efficient alternatives to standard attention layers~\citep[et cetera]{ katharopoulos2020transformers, bello2021lambdanetworks}. While they don't qualify as all-MLP models, they have close connections to our work (see Appendix~\ref{sec:app:lambda}) and aim at lowering the cost of AI, albeit it on fewer dimensions than our work (Appendix~\ref{sec:rw-greenai}). We employ FNet~\citep{lee-thorp2021fnetmixingtokens} and Linear Transformers~\citep{katharopoulos2020transformers} as representatives of these as a baseline.

}
{

\subsection{Green AI}\label{sec:rw-greenai}

\citet{schwartz2020green} challenges the current pursuit for higher accuracy at the cost of larger computation with the notion of "\green". Moreover, \citet{strubell2019energy} estimated the monetary and environmental cost of large model pretraining. Apart from being problematic environmentally, they argue that the monetary cost of pretraining is too high to be widely accessible for most researchers. In a research community that focuses on task performance, low resourced researchers would be disadvantaged. Therefore, metrics that take the cost of reaching a result are important to consider \citep{schwartz2020green}. The metric $Cost (R) \propto E \cdot D \cdot H$, is proposed and discussed in Section~\ref{sec:introduction}. However, reporting a single metric $Cost (R)$ is often ambiguous. Therefore, in our experiments, we consider the factors $E$, $D$, and $H$.

To measure the computational cost per example $E$, \citet{schwartz2020green} propose a count of the floating point operations (FPOs) required. In our experiments, we adopt this metric and further include wall-clock time for a practical application. The component $D$ evaluates the quantity of training data needed to reach a given accuracy or the performance of a model in a low-resource scenario~\citep{hedderich2020survey, chen2021empirical}. Finally, the component $H$ measures the cost associated with hyperparameter tuning. This is reported using \emph{expected validation performance} introduced by ~\citet{dodge2019show, dodge2021expected}, which computes the validation performance one would yield in expectation after $k$ hyperparameter trials of random search~\citep{bergstra2012random}.

Current literature does not focus on all facets of \green ~as formalized as $Cost(R)$. Typically, improving efficiency involves making existing models more accessible. For example, improving accessibility through model distillation~\citep{sanh2019distilbert} or adapter modules~\citep{houlsby2019parameter}.
Another avenue involves reducing the computational complexity, with examples: prompt-tuning~\citep{schick2020s}, self-attention in Transformers~\citep[et cetera]{child2019generating, beltagy2020longformer, katharopoulos2020transformers}. The latter approach is similar to our work. However, they focus the processing time of a single example $E$ and do not consider the other facets of \green. In our paper, we focus on MLP-based approaches, which we argue will have improvements in all facets of \green~due to their simplicity. 

\subsection{MLP-based Models}\label{sec:rw-allmlp}

The vision domain has seen promising results with purely MLP-based models ~\citep{tolstikhin2021mlp}, however, they lack the desired inductive biases for NLP. Some desirable properties for modeling language include: \textbf{i}) \emph{position invariance}, which is important for generalization, \textbf{ii}) \emph{adaptive size} for variable-length inputs, \textbf{iii}) a \emph{global receptive field}, which allows interactions to not be limited to small token neighborhoods, \textbf{iv}) \emph{learnabilty} allowing for universal applicablility to various tasks, and \textbf{v}) \emph{dynamicity} which implies that output is conditioned on the input.
MLP-based models are typically not used for NLP as including the inductive biases of position invariance, adaptive size and global receptive field are non-trivial for MLPs.

Several methods try to overcome the lack of adaptivity to size by introducing shifting operations and local windows. \citet{yu2021s2mlp} and \citet{lian2021asmlpan} uses spatial shifting to pass the information of adjacent tokens through an MLP. \citep{tang2021sparsemlpfor} uses a circular shifting operator. However, the position invariance is violated because positional information is required in the decision of which tokens are included in the neighborhood. The aggregation of local information itself is done via a (relative) position-specific MLP.
Global interactions are modeled only through the inclusion of enough layers or through a hierarchical layout~\citep{yu2021s2mlp,guo2021hiremlpvision}.

For vision tasks it can be useful to exploit the fact that 2D images consist of two axes. ~\citet{tatsunami2021raftmlphowmuch} make use of this fact by integrating a respective inductive bias. \cite{tu2022maxim} achieve linear complexity by applying a gMLP~\citep{liu2021payattentionto} to only a single axis.

A global receptive field in MLP-based models is achieved through token mixing and a weighted summation of the inputs, similar to self-attention. This allows for interaction between tokens. \citet{liu2021payattentionto} propose the model gMLP, where the mixing weights are determined by a fixed learnable interaction matrix between positions. However, this comes at the cost of violating position-invariance, size adaptivity, and dynamicity. DynaMixer~\citep{wang2022dynamixeravision} enables dynamicity by estimating the mixing weights from the concatenation of the inputs via a linear layer. This is efficient due to a dimensionality reduction step, but the concatenation still implies position-dependence and fixed-sized inputs. \cite{lee-thorp2021fnetmixingtokens} proposes the model FNet to use static Fourier transformations to model token interactions. This model made significant improvements in computation cost, although the functions lack learnability and are position dependent.

\subsection{Hypernetworks}

A hypernetwork uses a network to generate the weights for another, often larger, network \citep{ha2016hypernetworks}.
\citet{tay2020hypergridtransformers} leveraged task-conditioned hypernetworks for the GLUE benchmark. They achieved paralleled performance to the state-of-the-art at the time, whilst being more parameter efficient. \citet{mahabadi2021parameter} applied hypernetworks to Transformers to allow for parameter sharing in multitask learning. Their results showed parameter efficiencies and improved out of domain generation. \citet{zhmoginov2022hypertransformer} combine hypernetworks and transformers in the vision domain for few shot generalization.
LambdaNets are strongly related to our work, as they generate linear functions from context, in a similar capacity to a hypernetwork~\citep{bello2021lambdanetworks}. Their model is similar to the standard attention mechanism where the weights of three matrices $Q, K, V$ are learned. In contrast, HyperMixer uses the inputs to create non-linear transformations by generating an MLP. Features are combined based on their locations - a comparison can be found in Appendix~\ref{sec:app:lambda}.

Combining MLPMixer and hypernetworks allows for an efficient and simple MLP-based model to have  all the necessary inductive biases for NLP. The MLPMixer provides a simple token interaction backbone. By deploying hypernetworks to build the weights of the token mixing MLP, the missing inductive biases of position invariance and size adaptation are obtained.

}

\section{Experiments}
Our experiments are designed to test the following three hypotheses.
\hypothesisone~(Section~\ref{sec:results:peakperformance}): Since HyperMixer reflects more inductive biases that are adequate for NLP, our hypothesis is that HyperMixer performs better at NLP tasks than MLPMixer and similar MLP-based alternatives, specifically at those tasks that require to model the interactions between tokens.
\hypothesistwo: Since HyperMixer has similar inductive biases as transformers but is considerably simpler conceptually and in terms of computational complexity, it can be seen as a low cost alternative to Transformers, reducing the cost in terms of single example processing time (Section~\ref{sec:results:time-per-example}), required dataset size (Section~\ref{sec:results:low-resource-performance}), and hyperparameter tuning (Section~\ref{sec:results:ease-of-tuning}). 
\hypothesisthree~(Section~\ref{sec:results:locmix-attention}): Due to its inductive biases mirroring those of Transformers, HyperMixer also learns similar patterns as the attention mechanism.

\subsection{Datasets}
We evaluate on four sentence-pair classification tasks and one single-sentence classification task. The sentence-pair tasks are QQP \citep{WinNT}, QNLI \citep{rajpurkar2016squad}, MNLI \citep{williams2018broadcoverage} and SNLI \citep{bowman2015large}. For uniformity, datasets are formatted as in the GLUE benchmark~\citep{wang2018glue}.
We choose these tasks for two properties: firstly, they have large training datasets \short{(Table~\ref{tab:datasets}, appendix)}{(Table~\ref{tab:datasets})} enabling reasonable performances without pretraining;  secondly, solving these tasks requires good modeling of the interactions between tokens from different sentences, which is the main focus of this paper. As a control, we experiment on the single-input dataset SST2 \citep{socher2013recursive}, which is a sentiment classification task. Many examples in this dataset can be solved by identifying key sentiment words, rather than modeling the token interaction.

\short{}{
\begin{table}[h!]
    \centering
    \begin{tabular}{c|c|c|c}
    \textbf{Dataset} & \# \textbf{Train} & \# \textbf{Valid} & \# \textbf{Test} \\
    \hline
    MNLI & 392,702 & 9,815 & 9,796 \\
    SNLI & 549,367 & 9,842 & 9,824 \\
    QQP & 363,846 & 40,430 & 390,965 \\
    QNLI & 104,743 & 5,463 & 5,463 \\
    SST & 67,349 & 872 & 1,821 \\
    \hline
    \end{tabular}
    \caption{Number of examples in each dataset.}
    \label{tab:datasets}
\end{table}}

\subsection{Baselines}
The following baselines can be categorized into \emph{MLP-based} (to support \hypothesisone) and \emph{not MLP-based} (e.g., Transformers, to support \hypothesistwo). Note that our study is about the design of the \emph{token mixing} module.
Therefore, we only compare to models that fit into the general framework displayed in Figure~\ref{fig:model-architecture}, where there is a feature mixing module and a token mixing module for textual inputs. As a result, models such as RNNs are excluded. To enable a controlled experiment, we use the same feature mixing module in all models; the models only differ in their token mixing module.

\paragraph{MLP-based}

The conceptually closest baseline is \textbf{\texttt{MLPMixer}}~\citep{tolstikhin2021mlp},
which combines both token and feature mixing using fixed dimensional MLPs, as described in Section~\ref{sec:methods:mlpmixer}.
Concurrently, \cite{liu2021payattentionto} proposed \textbf{\texttt{gMLP}}, in which token mixing is achieved through weighted summation of all other inputs, similar to the attention mechanism. 
However, rather than computing weights as function of the inputs like in attention, in gMLP the weights are fixed learnable parameters. Additionally, linear gating initialized close to one is introduced to facilitate training. The original gMLP method does not employ feature mixing modules, as their token mixing module is capable of modeling feature interactions as well in a single gMLP block. However, for comparability we inject gMLP blocks as token mixing modules in our general architecture and keep feature mixing modules as well.

\paragraph{Non MLP-based}
\textbf{\texttt{Transformers}}~\citep{vaswani2017attention} are used in the current state of the art in virtually all NLP tasks. Their key component is the \emph{softmax}-based self-attention module, which we use for token mixing. 
\textbf{\texttt{Linear~Transformer}}~\citep{katharopoulos2020transformers} replaces softmax attention with a \textit{feature-map based dot-product attention}.
Finally, \textbf{\texttt{FNet}}~\citep{yu2021metaformerisactually} replaces the self-attention part of Transformers with a fixed, non-learnable set of Fourier transforms for token mixing.

\short{}
{

\short{We first describe the ablation models before we discuss their results.}{\subsubsection{Ablations}}

\paragraph{Feature Mixing Only}
The most simplistic MLP architecture is one that doesn't use token mixing, i.e., the token mixing module is set to the identity function.
The outputs at the last layer are aggregated via average pooling before plugged into the linear classifier.
This allows a baseline where the token interactions are not modeled. Therefore, this architecture serves as a control for how important token mixing is in any given task.

\paragraph{Token Mixing Only}
A simplistic single layer MLP architecture ablation. This model consists of a variable dimension MLP where the weights are generated using a hypernetwork which only allows for location interaction. This model is included to argue that the best simple model requires both location and feature mixing to efficiently model textual inputs.

\paragraph{Shared Weight-Vector}
A simple way to obtain a variable size location-mixing MLP is by weight-sharing. Concretely, we use a single learnable weight vector $w_1 \in \mathbb{R}^{d'}$, which we copy $N$ times to create a weight matrix $W_1 \in \mathbb{R}^{N \times d'}$. Analogously, we create $W_2$ from a separate vector $w_2$.
Note that this baseline does not support dynamicity, as the weight vector is independent of the inputs. This baseline thus shows the importance of dynamicity in our model.
}

\subsection{Performance}\label{sec:results:peakperformance}
Initially we compare the performance of HyperMixer in comparison to our baselines. Thereafter, we further explore the model's benefits with respects to its cost.

\short{For comparability, we adjust the size of the token mixing components such that all models have the same number of parameters (11M). FNet is an exception since it has no learnable parameters in its token mixing component. We tune the learning rate of each model via grid-search, and report the performance of the best configuration. Further experimental details on all experiments can be found in Appendix~\ref{sec:app-exp-details}.
}
{\paragraph{Experimental Setup}
To ensure a fair comparison, we aim to compare models of approximately the same number of parameters ($\approx$11 M parameters). All models have 6 layers with token embedding size $d = 256$ and hidden size $d' = 512$. For MLPMixer and gMLP we set the size of the token mixing modules to $N = 250$ and $N = 100$, respectively. These lengths are chosen to match the number of parameters of the other models (11 M). The hidden layer size is set to 512 in all models.
We use dropout at the input to each layer with a probability of 0.1.
For all models, including the ablations, we first tune the learning rate of Adam~\citep{kingma2014adam} using a logarithmically spaced grid of 7 values $\alpha \in \{0.001, 0.0005, 0.0002, 0.0001, 0.00005, 0.00002, \\ 0.00001\}$ on the validation set.
For our baselines, we then evaluate 10 different seeds and report the mean accuracy and standard deviation on the validation set.
On the test set, we only report the results of the model yielding the best results on the validation set, as the GLUE benchmark~\citep{wang2018glue} has a hidden test set with limited access.
Ablations are evaluated on the validation set with a single seed.
}

\paragraph{Results}
\short{Validation and test set results}{Ablations, validation and test set results} are shown in Table~\ref{tab:results-test}.
On the test and the validation set, HyperMixer performs the best among MLP-based models on all datasets, 
although for SST the difference on the validation set is smaller than one standard deviation.
MLPMixer generally achieves good performances, outperforming Transformers on two datasets.

Comparing to non-MLP-based methods, HyperMixer also outperforms vanilla Transformers on all datasets. The differences are generally small ($\leq 2$ points), except on QNLI, where the difference is 3.9 points. We suspect that this discrepancy is due to the relatively small training set of QNLI. We investigate low-resource behavior of Transformers in comparison to HyperMixer in Section~\ref{sec:results:low-resource-performance}.
FNet performs substantially worse than the other methods, particularly on SNLI and QQP. Linear Transformers achieve excellent performance on MNLI and SNLI, but perform poorly on QNLI and QQP.

\short{
In Appendix~\ref{sec:app-ablations}, we discuss ablations such as untied HyperMixer.}{
\short{Results are shown in Table~\ref{tab:ablations}.}{We now turn to the ablations.}
Untying the hypernetworks in HyperMixer leads to slightly decreased performance on all datasets.
We hypothesize that without pretraining, the model cannot benefits from more capacious token interaction modeling introduced by untying.
Nonetheless, the untied model still performs or a little better than vanilla Transformers.

While the introduction of MLPMixer and similar models follows a trend towards conceptually more simplistic models, our ablations show, perhaps unsurprisingly, that simplicity is not better when it leads to discarding information, as both the Feature-Mixing only and Location-Mixing only models perform substantially worse than the full HyperMixer model. Moreover, it is not enough to use the same learnable weight vector for all positions (Shared Weight-Vector), indicating the importance of generating the MLP based on the input.

The simplistic Feature-Mixing only model performs poorly on all datasets except SST, where it performs as well as the other models. This indicates that many instances in SST can be solved by looking at individual tokens alone, rather than modeling their interactions.
}

\begin{table*}
    \centering
    \begin{tabular}{c|c|c|c|c|c|r}
       \textbf{Model} & \textbf{MNLI} & \textbf{SNLI} & \textbf{QQP} & \textbf{QNLI} & \textbf{SST} & \# \textbf{Params} \\
        \hline
            \emph{Baselines} & \multicolumn{6}{c}{Validation set results (average accuracy / standard deviation over 10 seeds)}  \\
        \hline
        FNet & 59.7 (0.27) & 75.3 (0.46) & 79.4 (0.28) & 59.9 (0.46) & 79.7 (0.71) & 9.5 M \\
        Lin. Transformer &  \textbf{66.9} (0.48) & \textbf{82.7} (0.22) & 81.7 (0.28) & 61.3 (0.29) & 80.5 (0.46) & 11 M \\
        Transformer & 65.4 (0.51) & 80.9 (0.40) & 82.8 (0.22) & 67.3 (2.03) & 79.0 (0.86) & 11 M \\
        \hline
        MLPMixer & 63.9 (0.34) & 79.6 (0.11) & 83.7 (0.42) & 68.1 (2.10) & 80.1 (0.67) & 11 M \\
        gMLP & 60.8 (0.95) & 80.5 (0.55) & 82.8 (0.21) & 60.5 (0.49) & 78.7 (0.74) & 11 M \\
        HyperMixer (ours) &  \underline{66.2} (0.21) & \underline{81.9} (0.27) & \underline{\textbf{85.6}} (0.20) & \underline{\textbf{78.0}} (0.19) & 80.7 (0.84) & 11 M \\
        \hline
        \hline
        \short{}{
        \emph{Ablations} & \multicolumn{6}{c}{Validation set results (average accuracy / standard deviation over 10 seeds)}  \\
        \hline
        Feature Mixing only & 54.5 (0.25) & 67.0 (0.14) & 75.9 (0.06) & 60.8 (0.42) & 79.7 (0.64) & 9 M \\
        Token Mixing only & 59.0 (0.79) & 74.5 (5.53) & 79.5 (4.63) & 61.8 (1.29) & 76.3 (4.94) & 10 M \\
        Shared Weight-Vector & 57.1 (2.38) & 74.3 (1.96) & 82.9 (0.10) & 65.9 (0.42) & 79.8 (0.52) & 9.5 M \\
        HyperMixer (untied) & 65.8 (0.46) & 81.7 (0.30) & 84.8 (0.23) & 73.3 (0.53) & 80.3 (0.35) & 12 M \\
        \hline
                HyperMixer (tied) &  66.2 (0.21) & 81.9 (0.27) & 85.6 (0.20) & 78.0 (0.19) & 80.7 (0.84) & 11 M \\
        \hline
        }
        \emph{Baselines} & \multicolumn{6}{c}{Test set results (best model)} \\
        \hline
        FNet & 59.8 & 75.3 & 78.4 & 59.6 & 80.0 & 9.5 M \\
        Lin. Transformer & 66.9 & 83.0 & 82.3 & 61.7 & 80.8 & 11 M \\
        Transformer & 65.8 & 80.7 & 82.4 & 73.2 & 79.4 & 11 M \\
        \hline
        MLPMixer & 62.9 & 80.1 & 83.5 & 70.5 & 81.2 & 11 M \\
        gMLP & 61.2 & 80.9 & 82.5 & 60.2 & 79.5 & 11 M \\
        HyperMixer (ours) & 66.1 & 81.7 & 84.1 & 77.1 & 81.4 & 11 M \\
        \hline

    \end{tabular}
    \caption{\emph{Top}: Mean validation set accuracy and standard deviation over 10 different seeds of the best hyperparameter configuration. Results are printed in \textbf{bold} font if they exceed the second best result by at least one standard deviation. \underline{Underline} marks the best MLP-based model.
    \emph{Bottom}: Test set results on natural language understanding tasks when using the best model on the validation set. We evaluate on a single seed due to the limited test set access of GLUE.}
    \label{tab:results-test}
\end{table*}

\subsection{Time per Example}\label{sec:results:time-per-example}

In order to assess the efficiency of our model, we measure the wallclock-time of processing a single input (repeated 1,000 times) through the token mixing stages of HyperMixer and Transformer, respectively.
As \citet{schwartz2020green} point out, wallclock time has the downside of being dependent on the specific implementation, and they therefore recommend reporting the number of floating point operations (FOPs) required by one forward pass.\short{}{Due to the lack of reliable software to measure FOPs in PyTorch, we calculate these numbers manually. Our process is described in Appendix~\ref{sec:app:fops}.}
In Figure~\ref{fig:speed}, we show wallclock time and theoretical FOPs \short{}{of a single layer with $d = 256, d' = 512$ (as used in our experiments) }as a function of the input length $N$.
For short input sequences, the number of FOPs is dominated by the size of the hidden layer and hence slightly lower for Transformers than for HyperMixer. However, in practical terms we observe that HyperMixer is still faster than Transformers. At longer input sequences, the size of $N$ starts to dominate the total complexity of Transformers, so that it becomes exceedingly slower than HyperMixer.

\begin{figure}
    \centering
    \includegraphics[width=\columnwidth]{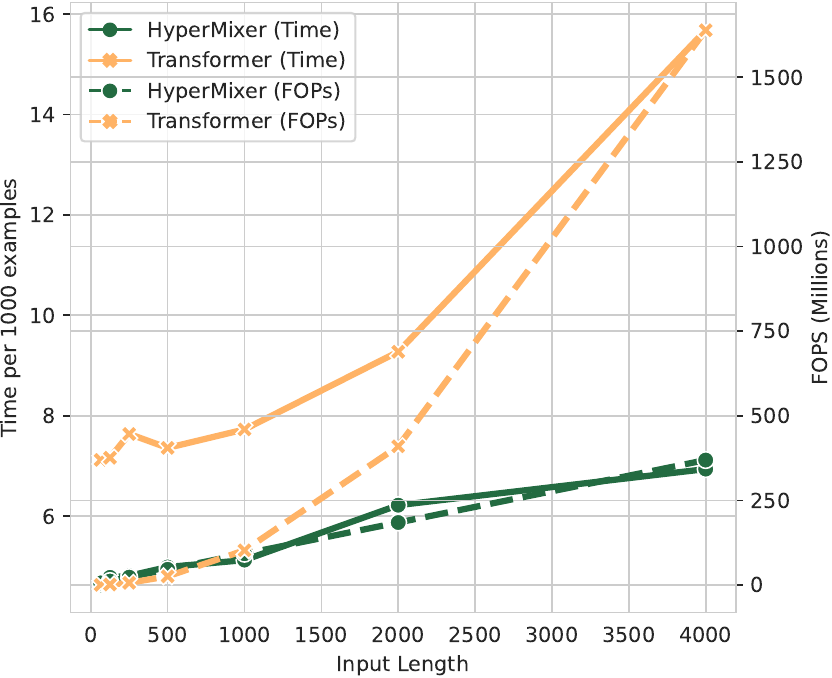}
    \caption{WCT / FOPs of propagating a single example through the token mixing of HyperMixer vs. Transformer depending on the input length.
    }
    \label{fig:speed}
\end{figure}

\subsection{Low Resource Performance}\label{sec:results:low-resource-performance}
\begin{figure}[t]
    \centering
    \includegraphics[width=\columnwidth]{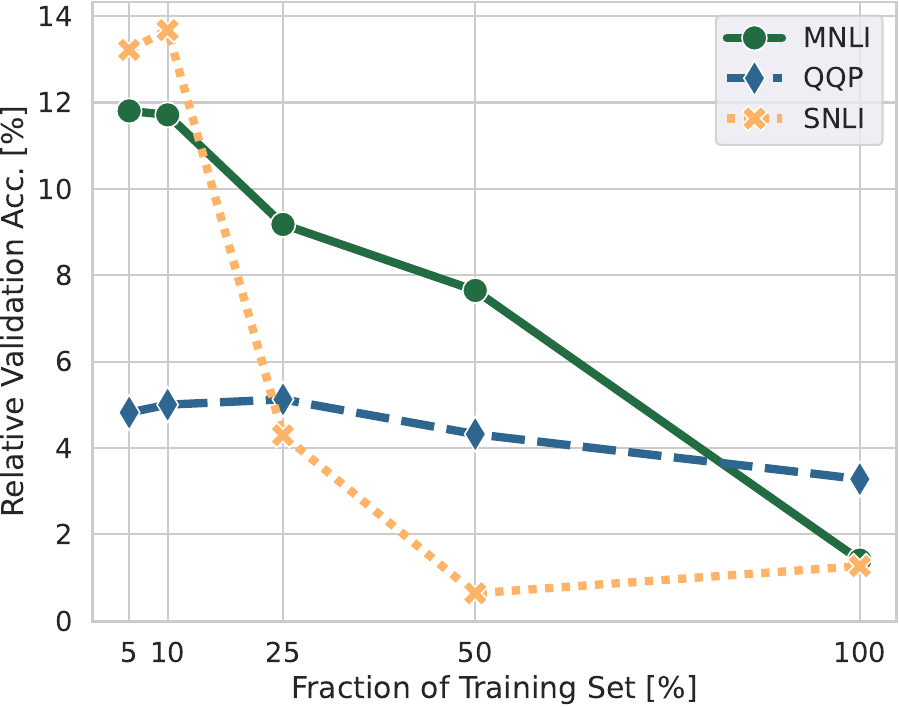}
    \caption{Relative improvement of HyperMixer over Transformer depending on what percentage of the training set is used.}
    \label{fig:low-resource}
\end{figure}

Like MLPMixer, HyperMixer is a conceptually simple architecture, as it only applies multi-layer perceptrons at its core. Simpler architectures often make for better performance on smaller scale datasets. We investigate  this by varying the number of examples used for training on the three large datasets MNLI, SNLI, and QQP.
For these experiments, we use the best performing learning rate found in the grid search from Section~\ref{sec:results:peakperformance}.
In Figure~\ref{fig:low-resource}, we plot the relative performance change of HyperMixer compared to Transformers as a function of subsample size.
On all datasets, the relative improvement of HyperMixer over Transformers is larger when training with 10\% of the dataset than with the full dataset.
While the effect is small on QQP, it is particularly large on SNLI and MNLI, where HyperMixer performs almost 12-14\% better with 10\% of the data, while the relative improvement with the full dataset is less than 2\%.

\subsection{Ease of Hyperparameter Tuning}\label{sec:results:ease-of-tuning}
MLP-based token mixing has the advantage that it is conceptually simpler than self-attention, and that it is well-known how to facilitate training via mechanisms such as skip-connections and layer normalization. Both these aspects suggest that it might be easier to find hyperparameter configurations that yield good performances.
In these experiments, we compare HyperMixer (with tied hypernetworks) to Transformers in this regard. As recommended in ~\citet{schwartz2020green}, we perform a random search to tune hyperparameters and compute the expected validation performance~\citep{dodge2019show, dodge2021expected}.
\short{Specifically, we tune the learning rate, whose logarithm is drawn from $\mathcal{U}(-8, -1)$, and the dropout probability drawn from $\mathcal{U}(0, 0.5)$ for 20 trials.}{
\paragraph{Experimental Setup}
In these experiments, we tune the initial learning rate of Adam and the dropout probability for regularization.
For the learning rate, its logarithm is drawn from $\mathcal{U}(-8, -1)$.
The dropout probability is drawn from $\mathcal{U}(0, 0.5)$.
}
\paragraph{Results}

\begin{figure}[t]
    \centering
    \includegraphics[width=\columnwidth]{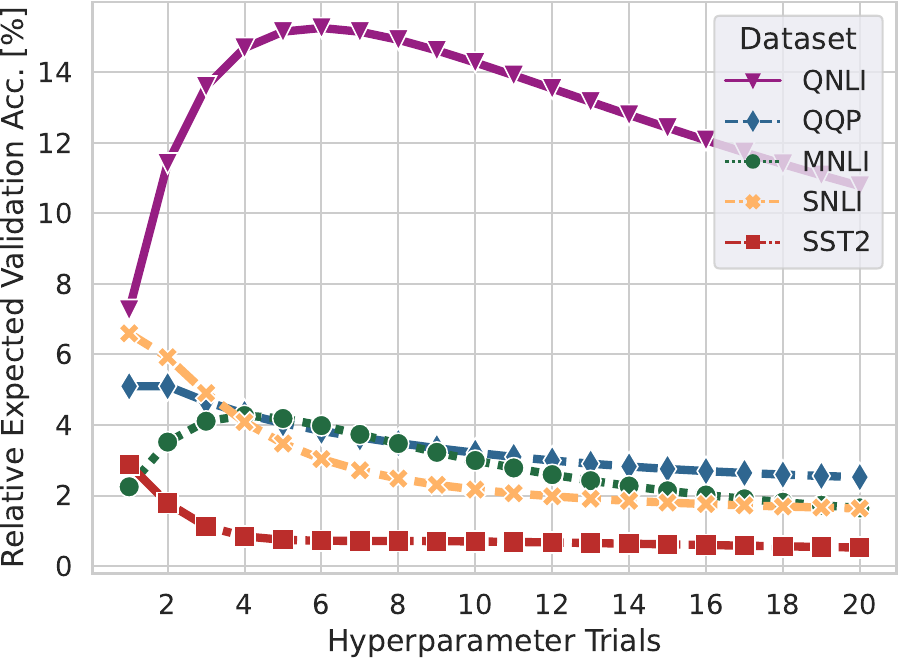}
    \caption{Relative expected validation performance of HyperMixer compared to Transformer after tuning the learning rate and dropout via random search.}
    \label{fig:tunability}
\end{figure}

In Figure~\ref{fig:tunability}, we show the \emph{relative} expected validation performance, i.e., the relative performance change of HyperMixer compared to Transformer, for all five datasets.
With the notable exception of QNLI, the relative improvement of HyperMixer is higher at smaller budgets than at larger budgets on all datasets. The effect is particularly strong on SNLI, where HyperMixer is 6.5\% better at small tuning budgets, but less than 2\% better at high budgets.
These results indicate that HyperMixer is substantially easier to tune than Transformers.

\subsection{HyperMixer Learns Attention Patterns}\label{sec:results:locmix-attention}

\begin{figure*}[t]
    \centering
    \begin{subfigure}[b]{0.5\textwidth}
         \centering
         \includegraphics[width=\textwidth]{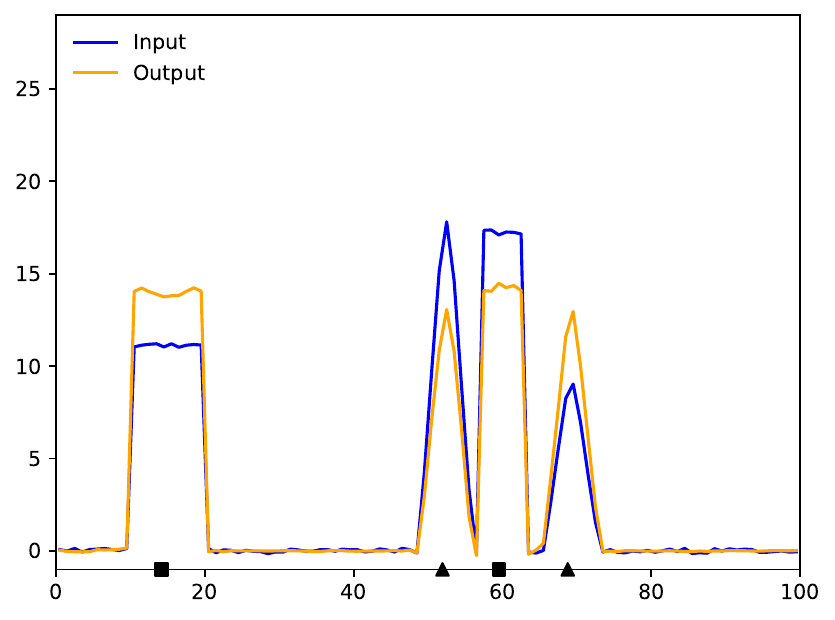}
         \caption{Example from the synthetic task}
         \label{fig:synth-example}
     \end{subfigure}\hfill
       \begin{subfigure}[b]{0.5\textwidth}
         \centering
         \includegraphics[width=\textwidth]{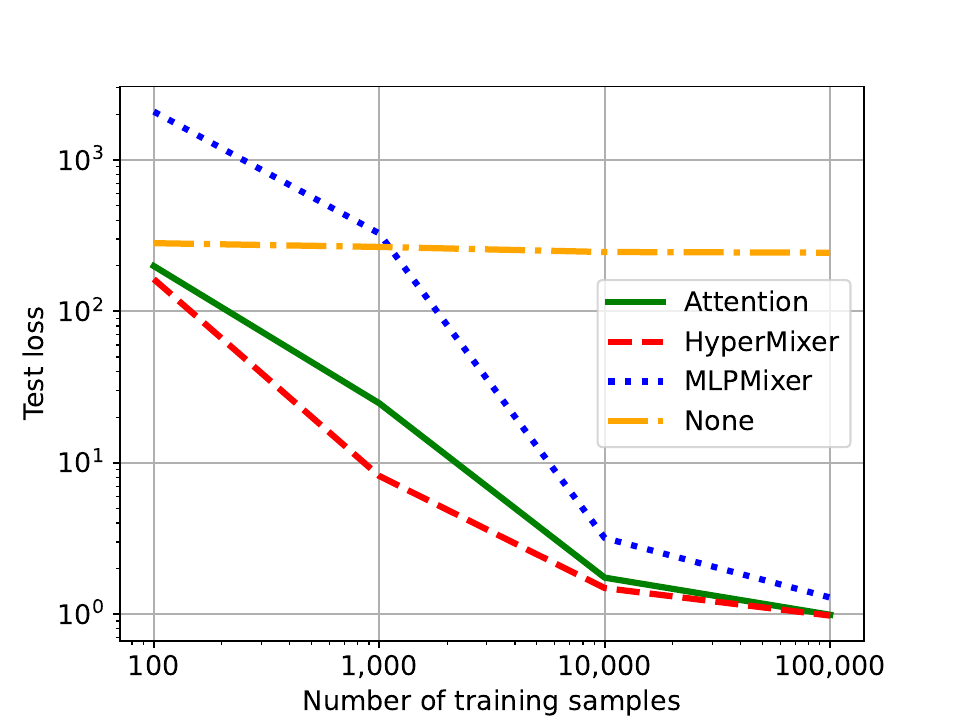}
         \caption{Test loss depending on number of examples.}
         \label{fig:toydata_test_loss}
     \end{subfigure}\hfill
     \short{}{
     \begin{subfigure}[b]{0.5\textwidth}
         \centering
        \includegraphics[width=\textwidth]{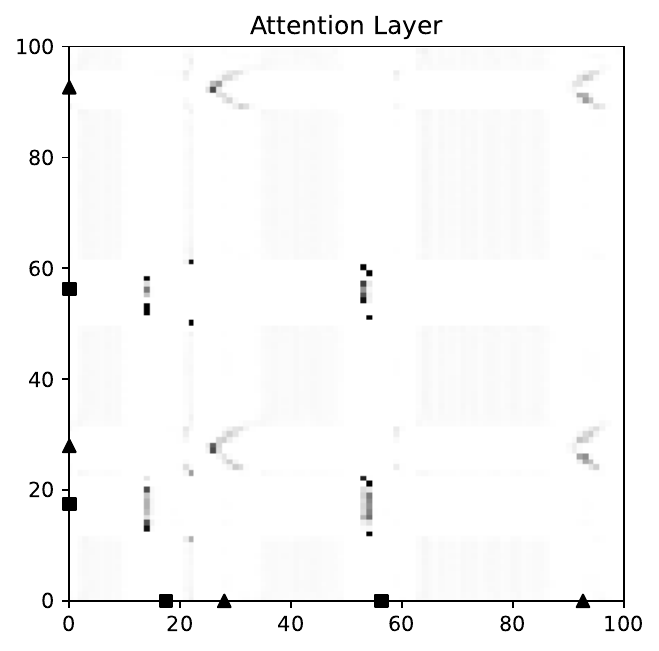}
        \caption{True attention map of Attention}
        \label{fig:wa_true}
     \end{subfigure}\hfill}
     \begin{subfigure}[b]{0.5\textwidth}
         \centering
         \includegraphics[width=\textwidth]{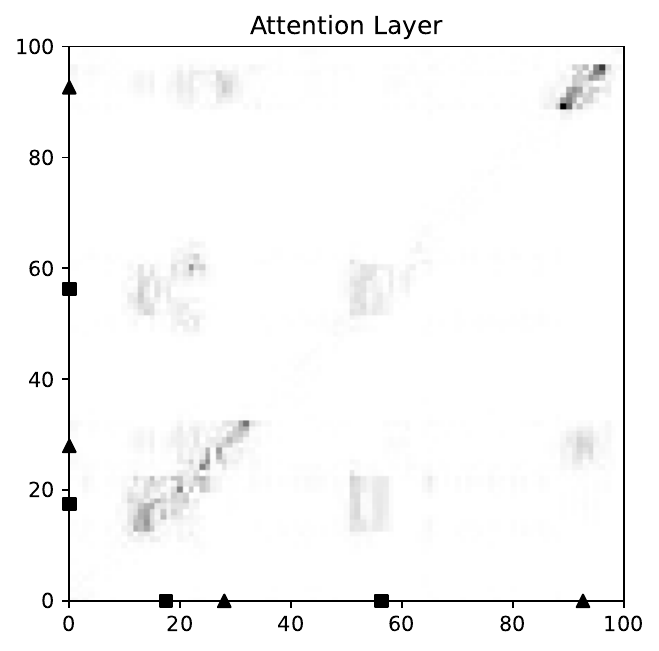}
         \caption{Pseudo-attention map of Attention}
         \label{fig:wa_pseudo}
     \end{subfigure}\short{}{\hfill
     \begin{subfigure}[b]{0.5\textwidth}
        \centering
         \includegraphics[width=\textwidth]{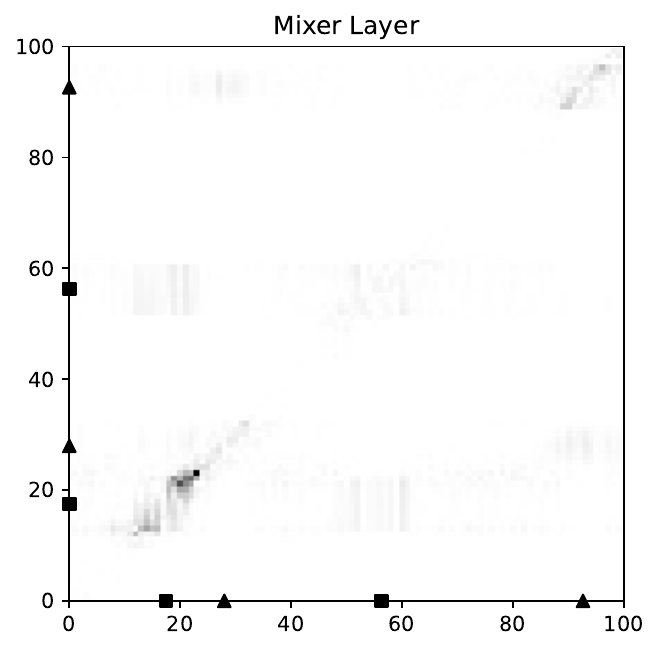}
         \caption{Pseudo-attention map of MLPMixer}
         \label{fig:mixer_pseudo}
     \end{subfigure}}\hfill
     \begin{subfigure}[b]{0.5\textwidth}
         \centering
         \includegraphics[width=\textwidth]{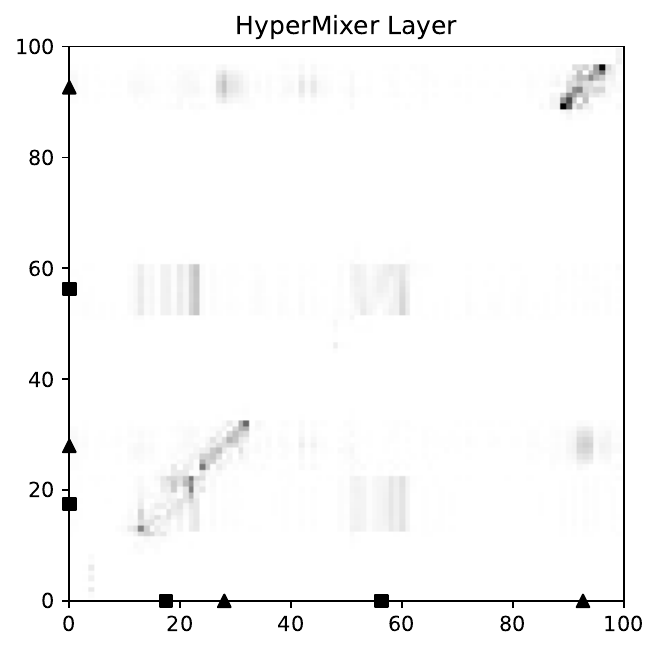}
         \caption{Pseudo-attention map of HyperMixer}
         \label{fig:hypermixer_pseudo}
     \end{subfigure}
     \caption{Results and \short{pseudo-}{(pseudo-)}attention maps on the synthetic task~\cite{fleuret2019attention}.}
     \label{fig:pseudo-att}
\end{figure*}

We hypothesized that the token mixing layer of HyperMixer offers a mechanism similar to attention. To show this, we consider a toy problem with 1d sequences composed of shape pairs of different heights as described in~\citet{fleuret2019attention}. The target value is the average height in each pair of shapes. An example input is shown in Figure~\ref{fig:synth-example}.
To solve the task well, for each position, the model must attend to other positions with the same shape.

\paragraph{Models} We compare the token mixing layer of HyperMixer to three other models: i) \emph{None} does not model token interactions. All predictions are thus only made based on local information. This model should thus fail. ii) \emph{MLPMixer} does model token interactions. Still, since its token mixing weights are position-specific, each position has to learn to recognize each shape, which we expect to be difficult, especially with little data. iii) \emph{Self-attention} can be considered the upper bound, as it models the interaction between every two positions explicitly.

\paragraph{Results}
Figure~\ref{fig:toydata_test_loss} shows the mean squared error on the test examples depending on the number of training examples.
As expected, \emph{None} fails on this task.
While all other models are able to solve the task with enough training data, MLPMixer is considerably less data-efficient than the other two models, requiring 5-10 times more data to reach the same performance. This is expected, since in contrast to HyperMixer and self-attention, MLPMixer's token mixing module is not position-invariant.
HyperMixer and self-attention reach approximately the same performance when training on 100k examples. However, HyperMixer is more data-efficient than self-attention, which we attribute to the simpler model architecture.

\short{}{\paragraph{Visualization of (pseudo-)attention weights}}
We can measure the interactions between two tokens by computing the gradient of an output token with respect to an input token (pseudo-attention).
Figures~\ref{fig:hypermixer_pseudo} and \ref{fig:wa_pseudo} show the pseudo-attention maps of \short{HyperMixer in comparison to attention.}{all models (trained on 25k examples) in comparison to the \emph{true} attention weights from the attention model.}
\short{We observe that the pseudo-attention weights of HyperMixer and attention are similar. This indicates that HyperMixer indeed learns an attention-like function. In contrast, we find these patterns to be weaker in MLPMixer (Figure~\ref{fig:pseudo-att-app}, appendix).
}
{First, it should be noted that pseudo-attention weights offer a somewhat blurry version of true attention weights, where high weights occur at positions that correspond to the same shape (cmp. \ref{fig:wa_true} to \ref{fig:wa_pseudo}).
Second, we observe that the pseudo-attention weights of HyperMixer and attention (cmp. Figure~\ref{fig:hypermixer_pseudo} to \ref{fig:wa_pseudo}) are similar. This indicates that HyperMixer indeed learns an attention-like transformation.
Third, MLPMixer also shows a similar pattern, but the relevant positions have weak connections. 
This confirms our finding that MLPMixer requires substantially more training data to learn strong connections.}

\section{Discussion}

In the following, we first discuss the merits of our proposed model, which are the core contributions of our paper.
We then discuss the scope of our analysis.

\subsection{Impact}
\paragraph{Best all-MLP model}
HyperMixer was designed as an MLP-based architecture with similar inductive biases as Transformers, which are beneficial for natural language understanding.
Our hypothesis (\hypothesisone) is that this leads to improvements over other MLP-based methods.
Our experimental results support this hypothesis, as we find HyperMixer to outperform all MLP-based baselines on all datasets (Section~\ref{sec:results:peakperformance}).

\paragraph{Low cost model}
The main motivation for an MLP-based architecture is the efficiency benefits induced by its simplicity.
Therefore, we hypothesized ($\hypothesistwo$) that HyperMixer would reduce the cost $Cost (R) \propto E \cdot D \cdot H$ to obtain an AI result $R$.
This hypothesis is supported by our experiments. While HyperMixer yields results that are on par with Transformer's results, it reduces the cost of all three cost factors: i) The cost of processing a single example (E) is lower, particularly for long inputs due to its linear complexity compared to the quadratic complexity of self-attention (Section~\ref{sec:results:time-per-example}). ii) The number of required training examples (D) is reduced, as HyperMixer's relative performance improvement is larger in the low-resource scenario (Section~\ref{sec:results:low-resource-performance}).
iii) HyperMixer requires less hyperparameter tuning than Transformers to reach good results, which is demonstrated by HyperMixer's higher expected relative improvements at low tuning budgets (Section~\ref{sec:results:ease-of-tuning}).

\paragraph{Attention-like model}
Finally, our experiments on a synthetic task indicate that HyperMixer can learn very similar attention patterns as the self-attention mechanism in Transformers (Section~\ref{sec:results:locmix-attention}), supporting hypothesis \hypothesisthree.
While MLPMixer can also learn similar patterns given enough training data, we believe that it is the introduction of adequate biases that allows HyperMixer to learn these patterns efficiently.
These biases were chosen based on an analysis of Transformer's success by \citet{henderson2020unstoppable}. HyperMixer's own success hence supports that analysis.

In summary, in our study, HyperMixer is the best-performing MLP-based architecture, and shows comparable performance and behavior as self-attention at substantially lower cost.
HyperMixer can thus be considered a low cost alternative to Transformers.

\subsection{Scope}\label{sec:discussion:limitations}
\paragraph{Small resource scenario}
It is important to note that our study is limited to the small resource scenario: Our models are small, not pretrained on large general-purpose corpora, and trained on datasets with fewer than 1 million examples.
It is unclear if our results will also hold on larger scale.
For example, while gMLP and FNet perform poorly in the low-resource scenario as demonstrated in our experiments, both models are able to narrow the gap to Transformer-based models as the resources for pretraining increase~\citep{liu2021payattentionto,lee-thorp2021fnetmixingtokens}.
We hypothesize that with enough resources, these models are able to overcome their shortcomings in terms of inductive biases.
However, there is no reason to believe that HyperMixer, being equipped with useful inductive biases, wouldn't perform on par with Transformers in high-resource scenarios while retaining its lower overall cost.
Quite the contrary, HyperMixer's linear complexity in sequence length perhaps makes it more appropriate for large-scale pretraining on long contexts than vanilla Transformers.

\paragraph{Versatility}
One of the most impressive qualities of Transformers is their versatility: Not only are they now the standard architecture for all NLP tasks, but over the years they have also become ubiquitous in a wide range of applications domains outside of NLP.
Of course, the present study cannot determine whether HyperMixer is as versatile as Transformers.
However, subsequent studies have shown that HyperMixer has uses in speech recognition~\citep{mai2023hyperconformer} and neural combinatorial optimization~\citep{drakulic2023bq}.
Still, some modeling advancements are needed. For example, HyperMixing is not yet applicable for decoder models that make use of causal masking. As decoder-only language models have become widely studied, this constitutes promising future work.

\section{Conclusion}
While large pretrained Transformer language models have led to impressive progress, they require so much resources that many research labs are excluded from participation, leading to calls for \emph{\green}.
We have proposed an MLP-based method, HyperMixer, that, in contrast to previous MLP-based methods, is equipped with the same inductive biases that made Transformers so successful for natural language understanding. While it performs on par with Transformers, it incurs substantially lower cost in terms of processing time, training data, and hyperparameter tuning.
Hence, we believe our study demonstrates the merits of MLP-based models for natural language understanding as an alternative to attention-based models, and we hope that the community pursues this direction further.
Avenues for future work include large-scale pretraining,
evaluation on a wider range of tasks and domains,
and the model’s adaptation to text generation.

\clearpage

\section*{Limitations}
Many limitations of our study are already discussed in Section~\ref{sec:discussion:limitations}, however, we repeat and add to them explicitly here.

\paragraph{Small resource scenario}
Our study investigates MLP-based architectures for text classification tasks and finds competitive performance with vanilla Transformers while having lower cost in terms of the Green AI equation.
However, the scope of our findings is naturally limited to the testing scenario, which is low-resource: Our models
are relatively small, not pretrained on large general-purpose corpora, and trained on datasets with fewer than 1 million examples. We may not say with certainty that our results will also hold on larger scale.
For the sake of hypothesis-driven research we consider it more valuable to run many controlled small-scale experiments rather than few large-scale experiments.
Nonetheless, scaling up should certainly be part of future research directions, as this is essential for optimal task performance.

\paragraph{Limitation to English pairwise sentence classification tasks}Since token mixing is the independent variable in our study, we put our main focus on English sentence-pair classification tasks with textual input only, which we presume (and provide some evidence for) to be most useful to assess differences between token mixing models.
Of course, vanilla Transformers are very flexible in the sense that, over the course of many studies, they have been shown to be very effective for a wide range of tasks, languages and data modalities.
Whether or not the proposed HyperMixer model possesses similar flexibility cannot be answered in this study. The HyperMixer encoder arguably possesses similar inductive biases as Transformers. We thus expect it to be straight-forward to apply to tasks that are also solved well by Transformer encoders (e.g., span classification).
For tasks such as language modeling, which involve a Transformer decoder, significant modeling advancements are required to obtain a HyperMixer equivalent. We consider this a very promising direction for future work.

\paragraph{Limitation to MLP-based baselines}
Similar to a trend in the computer vision community, our study investigates the suitability of MLP-based architectures for NLP. Due to their conceptual simplicity, these models promise to be easier to train, potentially leading to reduced Green AI costs.
To this end we compare our proposed HyperMixer model to a range of other MLP-based models, and Transformers.
Apart from FNet and Linear Transformers, which are efficient Transformer alternatives, we do not attempt an exhaustive comparison to non-MLP-based efficient NLP models.
Hence, the scope of our claims does not extend to all 
\emph{efficient} Transformer models.
However, these models are of course very relevant to this study, as they are targeted towards one of the factors of Green AI cost (single forward pass complexity). Therefore, we regard a comprehensive comparison as valuable future work.

\section*{Acknowledgements}
Florian Mai was supported by the Swiss National Science Foundation under the project
LAOS, grant
number 200021\_178862.
Arnaud Pannatier was supported by the Swiss Innovation Agency Innosuisse under the project MALAT, grant number ``32432.1 IP-ICT''. 
Fabio Fehr was supported by the Swiss National Centre of Competence in Research (NCCR) under the project Evolving Language, grant number ``51NF40\_180888''.
Haolin Chen was supported by the Swiss National Science Foundation under the project NAST, grant number ``185010''. 
François Marelli was supported by the Swiss National Science Foundation under the project COMPBIO, grant number ``179217''.

\bibliography{anthology,full_hypermixer}

\begin{thebibliography}{57}
\expandafter\ifx\csname natexlab\endcsname\relax\def\natexlab#1{#1}\fi

\bibitem[{Ba et~al.(2016)Ba, Kiros, and Hinton}]{ba2016layer}
Jimmy~Lei Ba, Jamie~Ryan Kiros, and Geoffrey~E Hinton. 2016.
\newblock Layer normalization.
\newblock \emph{arXiv preprint arXiv:1607.06450}.

\bibitem[{Bachlechner et~al.(2021)Bachlechner, Majumder, Mao, Cottrell, and
  McAuley}]{bachlechner2021rezero}
Thomas Bachlechner, Bodhisattwa~Prasad Majumder, Henry Mao, Gary Cottrell, and
  Julian McAuley. 2021.
\newblock Rezero is all you need: Fast convergence at large depth.
\newblock In \emph{Uncertainty in Artificial Intelligence}, pages 1352--1361.
  PMLR.

\bibitem[{Bello(2021)}]{bello2021lambdanetworks}
Irwan Bello. 2021.
\newblock \href {https://openreview.net/forum?id=xTJEN-ggl1b} {Lambdanetworks:
  Modeling long-range interactions without attention}.
\newblock In \emph{International Conference on Learning Representations}.

\bibitem[{Beltagy et~al.(2020)Beltagy, Peters, and
  Cohan}]{beltagy2020longformer}
Iz~Beltagy, Matthew~E Peters, and Arman Cohan. 2020.
\newblock Longformer: The long-document transformer.
\newblock \emph{arXiv preprint arXiv:2004.05150}.

\bibitem[{Bergstra and Bengio(2012)}]{bergstra2012random}
James Bergstra and Yoshua Bengio. 2012.
\newblock \href {http://jmlr.org/papers/v13/bergstra12a.html} {Random search
  for hyper-parameter optimization}.
\newblock \emph{Journal of Machine Learning Research}, 13(10):281--305.

\bibitem[{Bommasani et~al.(2021)Bommasani, Hudson, Adeli, Altman, Arora, von
  Arx, Bernstein, Bohg, Bosselut, Brunskill
  et~al.}]{bommasani2021opportunities}
Rishi Bommasani, Drew~A Hudson, Ehsan Adeli, Russ Altman, Simran Arora, Sydney
  von Arx, Michael~S Bernstein, Jeannette Bohg, Antoine Bosselut, Emma
  Brunskill, et~al. 2021.
\newblock On the opportunities and risks of foundation models.
\newblock \emph{arXiv preprint arXiv:2108.07258}.

\bibitem[{Bowman et~al.(2015)Bowman, Angeli, Potts, and
  Manning}]{bowman2015large}
Samuel~R. Bowman, Gabor Angeli, Christopher Potts, and Christopher~D. Manning.
  2015.
\newblock \href {https://doi.org/10.18653/v1/D15-1075} {A large annotated
  corpus for learning natural language inference}.
\newblock In \emph{Proceedings of the 2015 Conference on Empirical Methods in
  Natural Language Processing}, pages 632--642, Lisbon, Portugal. Association
  for Computational Linguistics.

\bibitem[{Chen et~al.(2021)Chen, Tam, Raffel, Bansal, and
  Yang}]{chen2021empirical}
Jiaao Chen, Derek Tam, Colin Raffel, Mohit Bansal, and Diyi Yang. 2021.
\newblock An empirical survey of data augmentation for limited data learning in
  nlp.
\newblock \emph{arXiv preprint arXiv:2106.07499}.

\bibitem[{Child et~al.(2019)Child, Gray, Radford, and
  Sutskever}]{child2019generating}
Rewon Child, Scott Gray, Alec Radford, and Ilya Sutskever. 2019.
\newblock Generating long sequences with sparse transformers.
\newblock \emph{arXiv preprint arXiv:1904.10509}.

\bibitem[{Chowdhery et~al.(2022)Chowdhery, Narang, Devlin, Bosma, Mishra,
  Roberts, Barham, Chung, Sutton, Gehrmann et~al.}]{chowdhery2022palm}
Aakanksha Chowdhery, Sharan Narang, Jacob Devlin, Maarten Bosma, Gaurav Mishra,
  Adam Roberts, Paul Barham, Hyung~Won Chung, Charles Sutton, Sebastian
  Gehrmann, et~al. 2022.
\newblock Palm: Scaling language modeling with pathways.
\newblock \emph{arXiv preprint arXiv:2204.02311}.

\bibitem[{Devlin et~al.(2019)Devlin, Chang, Lee, and
  Toutanova}]{devlin2018bert}
Jacob Devlin, Ming-Wei Chang, Kenton Lee, and Kristina Toutanova. 2019.
\newblock \href {https://doi.org/10.18653/v1/N19-1423} {{BERT}: Pre-training of
  deep bidirectional transformers for language understanding}.
\newblock In \emph{Proceedings of the 2019 Conference of the North {A}merican
  Chapter of the Association for Computational Linguistics: Human Language
  Technologies, Volume 1 (Long and Short Papers)}, pages 4171--4186,
  Minneapolis, Minnesota. Association for Computational Linguistics.

\bibitem[{Dodge et~al.(2019)Dodge, Gururangan, Card, Schwartz, and
  Smith}]{dodge2019show}
Jesse Dodge, Suchin Gururangan, Dallas Card, Roy Schwartz, and Noah~A. Smith.
  2019.
\newblock \href {https://doi.org/10.18653/v1/D19-1224} {Show your work:
  Improved reporting of experimental results}.
\newblock In \emph{Proceedings of the 2019 Conference on Empirical Methods in
  Natural Language Processing and the 9th International Joint Conference on
  Natural Language Processing (EMNLP-IJCNLP)}, pages 2185--2194, Hong Kong,
  China. Association for Computational Linguistics.

\bibitem[{Dodge et~al.(2021)Dodge, Gururangan, Card, Schwartz, and
  Smith}]{dodge2021expected}
Jesse Dodge, Suchin Gururangan, Dallas Card, Roy Schwartz, and Noah~A. Smith.
  2021.
\newblock \href {https://doi.org/10.18653/v1/2021.findings-emnlp.342} {Expected
  validation performance and estimation of a random variable{'}s maximum}.
\newblock In \emph{Findings of the Association for Computational Linguistics:
  EMNLP 2021}, pages 4066--4073, Punta Cana, Dominican Republic. Association
  for Computational Linguistics.

\bibitem[{Drakulic et~al.(2023)Drakulic, Michel, Mai, Sors, and
  Andreoli}]{drakulic2023bq}
Darko Drakulic, Sofia Michel, Florian Mai, Arnaud Sors, and Jean-Marc Andreoli.
  2023.
\newblock Bq-nco: Bisimulation quotienting for generalizable neural
  combinatorial optimization.
\newblock \emph{arXiv preprint arXiv:2301.03313}.

\bibitem[{Elman(1990)}]{elman1990finding}
Jeffrey~L Elman. 1990.
\newblock Finding structure in time.
\newblock \emph{Cognitive science}, 14(2):179--211.

\bibitem[{Fleuret(2019)}]{fleuret2019attention}
François Fleuret. 2019.
\newblock \href
  {https://fleuret.org/dlc/materials/dlc-slides-13-2-attention-mechanisms.pdf}
  {Attention mechanisms}.
\newblock Deep Learning Course - Chapter 13.2.

\bibitem[{Fodor and Pylyshyn(1988)}]{fodor1988connectionism}
Jerry~A Fodor and Zenon~W Pylyshyn. 1988.
\newblock Connectionism and cognitive architecture: A critical analysis.
\newblock \emph{Cognition}, 28(1-2):3--71.

\bibitem[{Fusco et~al.(2022)Fusco, Pascual, and Staar}]{fusco2022pnlp}
Francesco Fusco, Damian Pascual, and Peter Staar. 2022.
\newblock pnlp-mixer: an efficient all-mlp architecture for language.
\newblock \emph{arXiv preprint arXiv:2202.04350}.

\bibitem[{Guo et~al.(2021)Guo, Tang, Han, Chen, Wu, Xu, Xu, and
  Wang}]{guo2021hiremlpvision}
Jianyuan Guo, Yehui Tang, Kai Han, Xinghao Chen, Han Wu, Chao Xu, Chang Xu, and
  Yunhe Wang. 2021.
\newblock \href {https://arxiv.org/abs/2108.13341} {Hire-mlp: Vision mlp via
  hierarchical rearrangement}.

\bibitem[{Ha et~al.(2016)Ha, Dai, and Le}]{ha2016hypernetworks}
David Ha, Andrew Dai, and Quoc~V Le. 2016.
\newblock Hypernetworks.
\newblock \emph{arXiv preprint arXiv:1609.09106}.

\bibitem[{He et~al.(2016)He, Zhang, Ren, and Sun}]{he2016deep}
Kaiming He, Xiangyu Zhang, Shaoqing Ren, and Jian Sun. 2016.
\newblock Deep residual learning for image recognition.
\newblock In \emph{Proceedings of the IEEE conference on computer vision and
  pattern recognition}, pages 770--778.

\bibitem[{Hedderich et~al.(2020)Hedderich, Lange, Adel, Str{\"o}tgen, and
  Klakow}]{hedderich2020survey}
Michael~A Hedderich, Lukas Lange, Heike Adel, Jannik Str{\"o}tgen, and Dietrich
  Klakow. 2020.
\newblock A survey on recent approaches for natural language processing in
  low-resource scenarios.
\newblock \emph{arXiv preprint arXiv:2010.12309}.

\bibitem[{Henderson(2020)}]{henderson2020unstoppable}
James Henderson. 2020.
\newblock \href {https://doi.org/10.18653/v1/2020.acl-main.561} {The
  unstoppable rise of computational linguistics in deep learning}.
\newblock pages 6294--6306. Association for Computational Linguistics.

\bibitem[{Hendrycks and Gimpel(2016)}]{hendrycks2016gaussian}
Dan Hendrycks and Kevin Gimpel. 2016.
\newblock Gaussian error linear units (gelus).
\newblock \emph{arXiv preprint arXiv:1606.08415}.

\bibitem[{Houlsby et~al.(2019)Houlsby, Giurgiu, Jastrzebski, Morrone,
  De~Laroussilhe, Gesmundo, Attariyan, and Gelly}]{houlsby2019parameter}
Neil Houlsby, Andrei Giurgiu, Stanislaw Jastrzebski, Bruna Morrone, Quentin
  De~Laroussilhe, Andrea Gesmundo, Mona Attariyan, and Sylvain Gelly. 2019.
\newblock Parameter-efficient transfer learning for nlp.
\newblock In \emph{International Conference on Machine Learning}, pages
  2790--2799. PMLR.

\bibitem[{Iyer et~al.(2017)Iyer, Dandekar, and Csernai}]{WinNT}
Shankar Iyer, Nikhil Dandekar, and Kornel Csernai. 2017.
\newblock \href
  {https://quoradata.quora.com/First-Quora-Dataset-Release-Question-Pairs}
  {First quora dataset release: Question pairs}.

\bibitem[{Karimi~Mahabadi et~al.(2021)Karimi~Mahabadi, Ruder, Dehghani, and
  Henderson}]{mahabadi2021parameter}
Rabeeh Karimi~Mahabadi, Sebastian Ruder, Mostafa Dehghani, and James Henderson.
  2021.
\newblock \href {https://doi.org/10.18653/v1/2021.acl-long.47}
  {Parameter-efficient multi-task fine-tuning for transformers via shared
  hypernetworks}.
\newblock In \emph{Proceedings of the 59th Annual Meeting of the Association
  for Computational Linguistics and the 11th International Joint Conference on
  Natural Language Processing (Volume 1: Long Papers)}, pages 565--576, Online.
  Association for Computational Linguistics.

\bibitem[{Katharopoulos et~al.(2020)Katharopoulos, Vyas, Pappas, and
  Fleuret}]{katharopoulos2020transformers}
Angelos Katharopoulos, Apoorv Vyas, Nikolaos Pappas, and Fran{\c{c}}ois
  Fleuret. 2020.
\newblock Transformers are rnns: Fast autoregressive transformers with linear
  attention.
\newblock In \emph{International Conference on Machine Learning}, pages
  5156--5165. PMLR.

\bibitem[{Kingma and Ba(2014)}]{kingma2014adam}
Diederik~P Kingma and Jimmy Ba. 2014.
\newblock Adam: A method for stochastic optimization.
\newblock \emph{arXiv preprint arXiv:1412.6980}.

\bibitem[{Lee-Thorp et~al.(2021)Lee-Thorp, Ainslie, Eckstein, and
  Ontanon}]{lee-thorp2021fnetmixingtokens}
James Lee-Thorp, Joshua Ainslie, Ilya Eckstein, and Santiago Ontanon. 2021.
\newblock \href {https://arxiv.org/abs/2105.03824} {Fnet: Mixing tokens with
  fourier transforms}.

\bibitem[{Lhoest et~al.(2021)Lhoest, del Moral, Jernite, Thakur, von Platen,
  Patil, Chaumond, Drame, Plu, Tunstall et~al.}]{lhoest2021datasets}
Quentin Lhoest, Albert~Villanova del Moral, Yacine Jernite, Abhishek Thakur,
  Patrick von Platen, Suraj Patil, Julien Chaumond, Mariama Drame, Julien Plu,
  Lewis Tunstall, et~al. 2021.
\newblock Datasets: A community library for natural language processing.
\newblock \emph{arXiv preprint arXiv:2109.02846}.

\bibitem[{Lian et~al.(2022)Lian, Yu, Sun, and Gao}]{lian2021asmlpan}
Dongze Lian, Zehao Yu, Xing Sun, and Shenghua Gao. 2022.
\newblock \href {https://openreview.net/forum?id=fvLLcIYmXb} {{AS}-{MLP}: An
  axial shifted {MLP} architecture for vision}.
\newblock In \emph{International Conference on Learning Representations}.

\bibitem[{Liu et~al.(2021)Liu, Dai, So, and Le}]{liu2021payattentionto}
Hanxiao Liu, Zihang Dai, David So, and Quoc~V Le. 2021.
\newblock \href
  {https://proceedings.neurips.cc/paper/2021/file/4cc05b35c2f937c5bd9e7d41d3686fff-Paper.pdf}
  {Pay attention to mlps}.
\newblock In \emph{Advances in Neural Information Processing Systems},
  volume~34, pages 9204--9215. Curran Associates, Inc.

\bibitem[{Mai et~al.(2023)Mai, Zuluaga-Gomez, Parcollet, and
  Motlicek}]{mai2023hyperconformer}
Florian Mai, Juan Zuluaga-Gomez, Titouan Parcollet, and Petr Motlicek. 2023.
\newblock Hyperconformer: Multi-head hypermixer for efficient speech
  recognition.
\newblock In \emph{Interspeech 2023}. ISCA.

\bibitem[{Paszke et~al.(2019)Paszke, Gross, Massa, Lerer, Bradbury, Chanan,
  Killeen, Lin, Gimelshein, Antiga, Desmaison, Kopf, Yang, DeVito, Raison,
  Tejani, Chilamkurthy, Steiner, Fang, Bai, and Chintala}]{paszke2017automatic}
Adam Paszke, Sam Gross, Francisco Massa, Adam Lerer, James Bradbury, Gregory
  Chanan, Trevor Killeen, Zeming Lin, Natalia Gimelshein, Luca Antiga, Alban
  Desmaison, Andreas Kopf, Edward Yang, Zachary DeVito, Martin Raison, Alykhan
  Tejani, Sasank Chilamkurthy, Benoit Steiner, Lu~Fang, Junjie Bai, and Soumith
  Chintala. 2019.
\newblock \href
  {http://papers.neurips.cc/paper/9015-pytorch-an-imperative-style-high-performance-deep-learning-library.pdf}
  {Pytorch: An imperative style, high-performance deep learning library}.
\newblock In H.~Wallach, H.~Larochelle, A.~Beygelzimer, F.~d\'Alch\'{e} Buc,
  E.~Fox, and R.~Garnett, editors, \emph{Advances in Neural Information
  Processing Systems 32}, pages 8024--8035. Curran Associates, Inc.

\bibitem[{Radford et~al.(2018)Radford, Narasimhan, Salimans, and
  Sutskever}]{radford2018improving}
Alec Radford, Karthik Narasimhan, Tim Salimans, and Ilya Sutskever. 2018.
\newblock Improving language understanding by generative pre-training.

\bibitem[{Rajpurkar et~al.(2016)Rajpurkar, Zhang, Lopyrev, and
  Liang}]{rajpurkar2016squad}
Pranav Rajpurkar, Jian Zhang, Konstantin Lopyrev, and Percy Liang. 2016.
\newblock Squad: 100,000+ questions for machine comprehension of text.
\newblock In \emph{Proceedings of the 2016 Conference on Empirical Methods in
  Natural Language Processing}, pages 2383--2392.

\bibitem[{Sanh et~al.(2019)Sanh, Debut, Chaumond, and
  Wolf}]{sanh2019distilbert}
Victor Sanh, Lysandre Debut, Julien Chaumond, and Thomas Wolf. 2019.
\newblock Distilbert, a distilled version of bert: smaller, faster, cheaper and
  lighter.
\newblock \emph{arXiv preprint arXiv:1910.01108}.

\bibitem[{Schick and Sch{\"u}tze(2020)}]{schick2020s}
Timo Schick and Hinrich Sch{\"u}tze. 2020.
\newblock It's not just size that matters: Small language models are also
  few-shot learners.
\newblock \emph{arXiv preprint arXiv:2009.07118}.

\bibitem[{Schwartz et~al.(2020)Schwartz, Dodge, Smith, and
  Etzioni}]{schwartz2020green}
Roy Schwartz, Jesse Dodge, Noah~A Smith, and Oren Etzioni. 2020.
\newblock Green ai.
\newblock \emph{Communications of the ACM}, 63(12):54--63.

\bibitem[{Simonyan et~al.(2014)Simonyan, Vedaldi, and
  Zisserman}]{simonyan2014deep}
Karen Simonyan, Andrea Vedaldi, and Andrew Zisserman. 2014.
\newblock Deep inside convolutional networks: Visualising image classification
  models and saliency maps.
\newblock In \emph{In Workshop at International Conference on Learning
  Representations}. Citeseer.

\bibitem[{Socher et~al.(2013)Socher, Perelygin, Wu, Chuang, Manning, Ng, and
  Potts}]{socher2013recursive}
Richard Socher, Alex Perelygin, Jean Wu, Jason Chuang, Christopher~D Manning,
  Andrew~Y Ng, and Christopher Potts. 2013.
\newblock Recursive deep models for semantic compositionality over a sentiment
  treebank.
\newblock In \emph{Proceedings of the 2013 conference on empirical methods in
  natural language processing}, pages 1631--1642.

\bibitem[{Strubell et~al.(2019)Strubell, Ganesh, and
  McCallum}]{strubell2019energy}
Emma Strubell, Ananya Ganesh, and Andrew McCallum. 2019.
\newblock Energy and policy considerations for deep learning in nlp.
\newblock \emph{arXiv preprint arXiv:1906.02243}.

\bibitem[{Tang et~al.(2021)Tang, Zhao, Wang, Luo, Xie, and
  Zeng}]{tang2021sparsemlpfor}
Chuanxin Tang, Yucheng Zhao, Guangting Wang, Chong Luo, Wenxuan Xie, and Wenjun
  Zeng. 2021.
\newblock \href {https://arxiv.org/abs/2109.05422} {Sparse mlp for image
  recognition: Is self-attention really necessary?}

\bibitem[{Tatsunami and Taki(2021)}]{tatsunami2021raftmlphowmuch}
Yuki Tatsunami and Masato Taki. 2021.
\newblock \href {https://arxiv.org/abs/2108.04384} {Raftmlp: How much can be
  done without attention and with less spatial locality?}

\bibitem[{Tay et~al.(2021)Tay, Zhao, Bahri, Metzler, and
  Juan}]{tay2020hypergridtransformers}
Yi~Tay, Zhe Zhao, Dara Bahri, Don Metzler, and Da-Cheng Juan. 2021.
\newblock Hypergrid transformers: Towards a single model for multiple tasks.
\newblock In \emph{ICLR 2021}.

\bibitem[{Tolstikhin et~al.(2021)Tolstikhin, Houlsby, Kolesnikov, Beyer, Zhai,
  Unterthiner, Yung, Steiner, Keysers, Uszkoreit et~al.}]{tolstikhin2021mlp}
Ilya~O Tolstikhin, Neil Houlsby, Alexander Kolesnikov, Lucas Beyer, Xiaohua
  Zhai, Thomas Unterthiner, Jessica Yung, Andreas Steiner, Daniel Keysers,
  Jakob Uszkoreit, et~al. 2021.
\newblock Mlp-mixer: An all-mlp architecture for vision.
\newblock \emph{Advances in Neural Information Processing Systems}, 34.

\bibitem[{Tu et~al.(2022)Tu, Talebi, Zhang, Yang, Milanfar, Bovik, and
  Li}]{tu2022maxim}
Zhengzhong Tu, Hossein Talebi, Han Zhang, Feng Yang, Peyman Milanfar, Alan
  Bovik, and Yinxiao Li. 2022.
\newblock Maxim: Multi-axis mlp for image processing.
\newblock \emph{arXiv preprint arXiv:2201.02973}.

\bibitem[{Vaswani et~al.(2017)Vaswani, Shazeer, Parmar, Uszkoreit, Jones,
  Gomez, Kaiser, and Polosukhin}]{vaswani2017attention}
Ashish Vaswani, Noam Shazeer, Niki Parmar, Jakob Uszkoreit, Llion Jones,
  Aidan~N Gomez, {\L}ukasz Kaiser, and Illia Polosukhin. 2017.
\newblock Attention is all you need.
\newblock \emph{Advances in neural information processing systems}, 30.

\bibitem[{Wang et~al.(2018)Wang, Singh, Michael, Hill, Levy, and
  Bowman}]{wang2018glue}
Alex Wang, Amanpreet Singh, Julian Michael, Felix Hill, Omer Levy, and Samuel~R
  Bowman. 2018.
\newblock Glue: A multi-task benchmark and analysis platform for natural
  language understanding.
\newblock \emph{arXiv preprint arXiv:1804.07461}.

\bibitem[{Wang and Komatsuzaki(2021)}]{wang2021gpt}
Ben Wang and Aran Komatsuzaki. 2021.
\newblock Gpt-j-6b: A 6 billion parameter autoregressive language model.

\bibitem[{Wang et~al.(2019)Wang, Li, Xiao, Zhu, Li, Wong, and
  Chao}]{wang2019learning}
Qiang Wang, Bei Li, Tong Xiao, Jingbo Zhu, Changliang Li, Derek~F Wong, and
  Lidia~S Chao. 2019.
\newblock Learning deep transformer models for machine translation.
\newblock In \emph{Proceedings of the 57th Annual Meeting of the Association
  for Computational Linguistics}, pages 1810--1822.

\bibitem[{Wang et~al.(2022)Wang, Jiang, Zhu, Yuan, Song, and
  Liu}]{wang2022dynamixeravision}
Ziyu Wang, Wenhao Jiang, Yiming Zhu, Li~Yuan, Yibing Song, and Wei Liu. 2022.
\newblock \href {https://arxiv.org/abs/2201.12083} {Dynamixer: A vision mlp
  architecture with dynamic mixing}.

\bibitem[{Williams et~al.(2018)Williams, Nangia, and
  Bowman}]{williams2018broadcoverage}
Adina Williams, Nikita Nangia, and Samuel Bowman. 2018.
\newblock \href {https://doi.org/10.18653/v1/N18-1101} {A broad-coverage
  challenge corpus for sentence understanding through inference}.
\newblock In \emph{Proceedings of the 2018 Conference of the North {A}merican
  Chapter of the Association for Computational Linguistics: Human Language
  Technologies, Volume 1 (Long Papers)}, pages 1112--1122, New Orleans,
  Louisiana. Association for Computational Linguistics.

\bibitem[{Yu et~al.(2022)Yu, Li, Cai, Sun, and Li}]{yu2021s2mlp}
Tan Yu, Xu~Li, Yunfeng Cai, Mingming Sun, and Ping Li. 2022.
\newblock S2-mlp: Spatial-shift mlp architecture for vision.
\newblock In \emph{Proceedings of the IEEE/CVF Winter Conference on
  Applications of Computer Vision (WACV)}, pages 297--306.

\bibitem[{Yu et~al.(2021)Yu, Luo, Zhou, Si, Zhou, Wang, Feng, and
  Yan}]{yu2021metaformerisactually}
Weihao Yu, Mi~Luo, Pan Zhou, Chenyang Si, Yichen Zhou, Xinchao Wang, Jiashi
  Feng, and Shuicheng Yan. 2021.
\newblock \href {https://arxiv.org/abs/2111.11418} {Metaformer is actually what
  you need for vision}.

\bibitem[{Zhmoginov et~al.(2022)Zhmoginov, Sandler, and
  Vladymyrov}]{zhmoginov2022hypertransformer}
Andrey Zhmoginov, Mark Sandler, and Max Vladymyrov. 2022.
\newblock Hypertransformer: Model generation for supervised and semi-supervised
  few-shot learning.
\newblock \emph{arXiv preprint arXiv:2201.04182}.

\end{thebibliography}
\bibliographystyle{acl_natbib}

\cleardoublepage
\appendix

{\bf{\Huge{Appendix}}}

\section{Extended Related Work}

\section{Experimental Details}\label{sec:app-exp-details}

\subsection{General Information}

\paragraph{Implementation}
We implemented all models within the same general framework based on PyTorch~\citep{paszke2017automatic}. An implementation of the layer is available at \href{https://github.com/idiap/hypermixing}{https://github.com/idiap/hypermixing}. For tokenization, we use the pretrained tokenizer from BERT-Base~\citep{devlin2018bert}. Datasets are downloaded directly from HuggingFace Datasets~\citep{lhoest2021datasets}. As such, they are directly downloaded by our training script. We apply no further preprocessing.

For computing expected validation performance, we use the public implementation by \citet{dodge2019show}.

We run our experiments on single-GPU servers available to us as part of a computation grid, ranging between GeForce GTX Titan X and RTX 3090. Apart from Transformers on SNLI and MNLI, which take about 4 hours on slower GPUs, all experiments finished within 3 hours. 
\paragraph{Hyperparameters} We provide CSV files detailing all parameters of every run alongside their results in the supplementary material, ensuring reproducibility of our study. Note that the computation environment (e.g., type of GPU) might lead to small differences.

\subsection{Peak Performance}
To ensure a fair comparison, we aim to compare models of approximately the same number of parameters ($\approx$11 M parameters). All models have 6 layers with token embedding size $d = 256$ and hidden size $d' = 512$. For MLPMixer and gMLP we set the size of the token mixing modules to $N = 250$ and $N = 100$, respectively. These lengths are chosen to match the number of parameters of the other models (11 M). The hidden layer size is set to 512 in all models.
We use dropout at the input to each layer with a probability of 0.1.
For all models, including the ablations, we first tune the learning rate of Adam~\citep{kingma2014adam} using a logarithmically spaced grid of 7 values $\alpha \in \{0.001, 0.0005, 0.0002, 0.0001, 0.00005, 0.00002, \\ 0.00001\}$ on the validation set.
For our baselines, we then evaluate 10 different seeds and report the mean accuracy and standard deviation on the validation set.
On the test set, we only report the results of the model yielding the best results on the validation set, as the GLUE benchmark~\citep{wang2018glue} has a hidden test set with limited access.
Ablations are evaluated on the validation set with a single seed.

\short{\begin{table}[h!]
    \centering
    \begin{tabular}{c|c|c|c}
    \textbf{Dataset} & \# \textbf{Train} & \# \textbf{Valid} & \# \textbf{Test} \\
    \hline
    MNLI & 392,702 & 9,815 & 9,796 \\
    SNLI & 549,367 & 9,842 & 9,824 \\
    QQP & 363,846 & 40,430 & 390,965 \\
    QNLI & 104,743 & 5,463 & 5,463 \\
    SST & 67,349 & 872 & 1,821 \\
    \hline
    \end{tabular}
    \caption{Number of examples in each dataset.}
    \label{tab:datasets}
\end{table}}{}

\subsection{Time per Example}
Due to the lack of reliable software to measure FOPs in PyTorch, we calculate these numbers manually. Our process is described in Appendix~\ref{sec:app:fops}.
For the measurement of wallclock time, we measured the time of 1,000 batches through a single layer of each token mixing module with $d = 256, d' = 512$ (as used in our experiments).

\subsection{Toy Task (Section~\ref{sec:results:locmix-attention})}
This section gives more detail about how we set up the synthetic example \cite{fleuret2019attention} for evaluating whether the different models were able to learn some attention-like transformation. 
We have a dataset made of 1D sequences that contain two rectangular and two triangular shapes. Each of these shapes has a different height taken at random in the input sequence. The output sequence has the same shapes in the same positions, but the heights of triangular shapes should be the mean of the two triangular shapes in the input sequence. Similarly, the height of the rectangular shapes in the output sequence is the mean of the height of the two rectangular shapes in the input sequence.  

So the model should be able to see across the sequence and compute the mean of the two different shapes to succeed at the task. All the models considered for this task have a similar structure: they consist of a particular layer (MLPMixer, HyperMixer, or Attention) surrounded by two pairs of 1D-convolutional layers with kernels of size five and a symmetric zero-padding of size two so that the output shape is constant. We made an ablation to ensure that this layer was mandatory by changing it with another similar 1D convolutional layer, which corresponds to None in the figure \ref{fig:toydata_test_loss}. 

Before visualizing the pseudo-attention maps, all models were trained on 25,000 training examples.
We use input-gradients \cite{simonyan2014deep} to evaluate whether models could « attend » to the different shapes. This method computes the gradient of the output sequence with respect to the input sequence, giving the corresponding saliency map, which can then be recombined into a pseudo-attention matrix where the $i$-th column corresponds to the saliency maps of the $i$-th output token. A large value in the $(i,j)$ entries of the pseudo-attention matrix means that the output token $i$ strongly depends on the input $j$, and we can thus compare it to an attention matrix \ref{fig:wa_true-app}. 

Figure \ref{fig:pseudo-att-app} represents the pseudo-attention matrices for the different models. We can notice that it indeed approximates the true attention matrix \ref{fig:wa_true-app} and that the model with no special layer cannot attend to the correct part of the sequence, as expected. Finally, we can see that the pseudo-attention of the Mixer layer is not as peaked as the one corresponding to the Attention or HyperMixer layer.

\section{Further Results}

\subsection{Validation Set Results}

In Table~\ref{tab:results-dev}, we show the best scores on the validation set that we obtained from the grid search (using a fixed seed), alongside the learning rate that yielded that score.
\begin{table*}
    \centering
    \resizebox{\textwidth}{!}{\begin{tabular}{c|c|c|c|c|c|r}
    \hline
       \textbf{Model} & \textbf{MNLI} & \textbf{SNLI} & \textbf{QQP} & \textbf{QNLI} & \textbf{SST} & \# \textbf{Params} \\
        \hline
             & \multicolumn{6}{c}{Baselines (accuracy / learning rate)}  \\
        \hline
        FNet & 59.6 / 5e-4 & 75.1 / .001 & 79.7 / .001 & 59.2 / 5e-4 & 80.4 / .001 & 9.5 M \\
        Linear Transformer & 66.2 / .001 & 82.2 / 0.001 & 81.7 / 5e-4 & 61.1 / 1e-4 & 80.7 / 2e-4 & 11M \\
        Transformer & 66.0 / 2e-4 & 81.2 / 2e-4 & 82.9 / 2e-4 & 65.4 / 5e-4 & 78.9 / 5e-4 & 11 M \\
        \hline
        MLPMixer & 64.2 / .001 & 80.5 / .001 & 83.6 / .001 & 68.7 / 5e-5 & 82.3 / .001 & 11 M \\
        gMLP & 61.5 / .001 & 80.9 / 2e-4 & 83.0 / 5e-4 & 61.1 / 5e-5 & 79.2 / 1e-4 & 11 M \\
        \hline
        \hline
        HyperMixer (tied) & 66.5 / 1e-4 & 81.8 / 2e-4 & 85.4 / 1e-4 & 77.5 / 5e-5 & 81.3 / 5e-4 & 11 M \\
        \hline
        & \multicolumn{6}{c}{Ablations (accuracy / learning rate)} \\
        \hline
        Feature Mixing only & 54.4 / .001 & 67.2 / 5e-4 & 75.9 / .001 & 61.0 / .001 & 81.8 / 5e-4 & 9 M \\
        Token Mixing only & 59.5 / 2e-4 & 73.6 / 2e-4 & 81.7 / 2e-4 & 61.8 / 2e-4 & 80.1 / 5e-4 & 9 M \\
        Shared Weight-Vector & 53.7 / 5e-4 & 68.1 / .001 & 83.0 / .001 & 66.4 / 5e-5 & 80.5 / .001 & 9.5 M \\
        HyperMixer (untied) & 66.0 / .001 & 82.3 / .001 & 84.6 / .001 & 72.2 / 5e-5 & 81.3 / .001 & 12 M \\
        \hline
    \end{tabular}}
    \caption{Best validation set results on natural language understanding tasks after tuning the learning rate on a grid.}
    \label{tab:results-dev}
\end{table*}

In Section~\ref{sec:results:peakperformance}, we reported the test set results of all models when using the best-performing seed. In Table~\ref{tab:results-test-median}, we show test set results when using the median seed.

\begin{table*}
    \centering
    \begin{tabular}{c|c|c|c|c|c|r}
    \hline
       \textbf{Model} & \textbf{MNLI} & \textbf{SNLI} & \textbf{QQP} & \textbf{QNLI} & \textbf{SST} & \# \textbf{Params} \\
        \hline
             & \multicolumn{6}{c}{Baselines} \\
        \hline
        FNet & 58.8 & 75.2 & 78.4 & 59.0 & 80.2 & 9.5 M \\
        Lin. Transformer & 67.0 & 81.9 & 82.3 & 61.0 & 82.5 & 11 M \\
        Transformer & 64.9 & 81.1 & 82.1 & 67.1 & 77.7 & 11 M \\
        \hline
        MLPMixer & 62.6 & 79.7 & 83.2 & 69.1 & 80.8 & 11 M \\
        gMLP & 62.9 & 79.9 & 82.3 & 60.0 & 78.5 & 11 M \\
        \hline
        \hline
        HyperMixer (tied) & 64.9 & 81.0 & 83.9 & 76.8 & 80.9 & 11 M \\
        \hline
    \end{tabular}
    \caption{Test set results on natural language understanding tasks, when using the median seed.}
    \label{tab:results-test-median}
\end{table*}

\short{
\subsection{Ablations}\label{sec:app-ablations}

\paragraph{Results}

\begin{table*}
    \centering
    \resizebox{\textwidth}{!}{
    \begin{tabular}{c|c|c|c|c|c|r}
       \textbf{Model} & \textbf{MNLI} & \textbf{SNLI} & \textbf{QQP} & \textbf{QNLI} & \textbf{SST} & \# \textbf{Params} \\
        \hline
        
    \end{tabular}}
    \caption{Mean and standard deviation of HyperMixer ablations on the validation set.}
    \label{tab:ablations}
\end{table*}
}{}

\short{
\subsection{Visualizing Attention Patterns}
Figure~\ref{fig:pseudo-att-app} shows the pseudo-attention of all models (except 'None') alongside the true attention weights of attention.
First, it should be noted that pseudo-attention weights offer a somewhat blurry version of true attention weights, where high weights occur at positions that correspond to the same shape (cmp. \ref{fig:wa_true-app} to \ref{fig:wa_pseudo-app}).
Second, we observe that the pseudo-attention weights of HyperMixer and attention (cmp. Figure~\ref{fig:hypermixer_pseudo-app} to \ref{fig:wa_pseudo-app}) are similar. This indicates that HyperMixer indeed learns an attention-like function.
Third, MLPMixer also shows a similar pattern, but the relevant positions have weak connections (Figure~\ref{fig:mixer_pseudo-app}). 
This confirms our finding that MLPMixer requires substantially more training data to learn strong connections.
\begin{figure*}
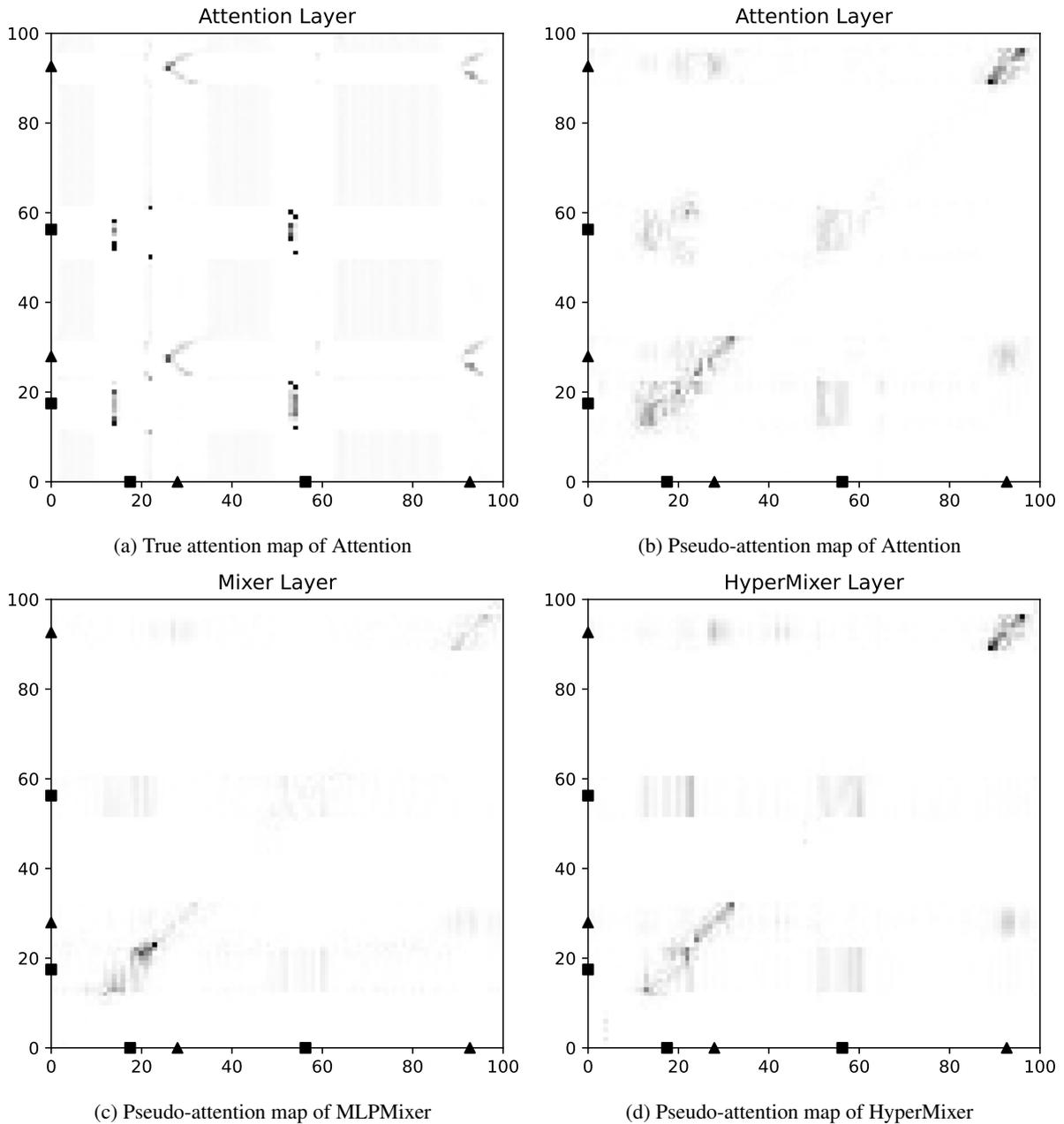

    \centering
     \begin{subfigure}[b]{0.5\textwidth}
         \centering
        \includegraphics[width=\textwidth]{pseudo-attention/att1d_wa_test_true_A_000.pdf}
        \caption{True attention map of Attention}
        \label{fig:wa_true-app}
     \end{subfigure}\hfill
     \begin{subfigure}[b]{0.5\textwidth}
         \centering
         \includegraphics[width=\textwidth]{pseudo-attention/attention-pseudo-attention.pdf}
         \caption{Pseudo-attention map of Attention}
         \label{fig:wa_pseudo-app}
     \end{subfigure}\hfill
     \begin{subfigure}[b]{0.5\textwidth}
        \centering
         \includegraphics[width=\textwidth]{pseudo-attention/mixer-pseudo-attention.pdf}
         \caption{Pseudo-attention map of MLPMixer}
         \label{fig:mixer_pseudo-app}
     \end{subfigure}\hfill
     \begin{subfigure}[b]{0.5\textwidth}
         \centering
         \includegraphics[width=\textwidth]{pseudo-attention/hypermixer-pseudo-attention.pdf}
         \caption{Pseudo-attention map of HyperMixer}
         \label{fig:hypermixer_pseudo-app}
     \end{subfigure}
     \caption{Results and (pseudo-)attention maps on the synthetic task~\cite{fleuret2019attention}.}
     \label{fig:pseudo-att-app}
\end{figure*}
}{}

\section{Comparison of \#FOP}\label{sec:app:fops}

We want to compute the number of floating-point operations needed in self-attention vs. HyperMixing for a single example. Let $N$ be the sequence length, $d$ be the embedding size of each token, and $d'$ the hidden dimension.

For simplicity, we will assume basic mathematical operators like $\exp, \tanh, \sqrt{x}$ and division to be equal to one floating operation. However, their actual cost is higher but depends on implementation and hardware.

\subsection{Basic Building Blocks}
We first compute the number of operations infrequently occurring in basic building blocks of neural networks.

\paragraph{Matrix Multiplication}
Multiplying matrix $A \in \mathbb{R}^{N\times d}$ $A \in \mathbb{R}^{d \times M}$ takes $2d(NM)$ operations, as $2d$ operations are needed for a single dot-product and there are $NM$ entries in the resulting matrix.

\paragraph{Linear Layer}
Passing a single vector of size $d$ through a linear layer without bias of size $(d, d')$ is the multiplication of a single vector with a matrix, i.e., incurs $2dd'$ operations in total.

\paragraph{GELU}
GELU is usually approximated as
\begin{equation*}
\operatorname{GELU}(x) = 0.5x \left[ 1  + \operatorname{tanh}\left(\sqrt{2/\pi}(x + cx^3) \right)\right]
\end{equation*}
So in total, GELU is computed for every of the $d$ features and every of the $N$ vectors, meaning the GELU activation layer takes $9dN$ operations.

\paragraph{MLP (input = output size)}
Given hidden size $d'$ and input/output size $d$, we have two linear layers of size $(d, d')$ and $(d', d)$, respectively, plus a GELU layer on $d'$ dimensions, incurring $4dd' + 9d'$.

\paragraph{MLP (input /= output size)}
Given hidden size $d'$, input size $d$ and output size $d''$, we have two linear layers of sizes $(d, d')$ and $(d', d'')$, incurring $2dd'$ + $2d'd'' + 9d'$.

\paragraph{Softmax}
Softmax is applied over $N$ values, each of which goes through an $\exp$ and a division by the normalization value. The normalization value requires $N$ additions.
So in total, the number of operations is $3N$.

\subsection{HyperMixer}

\paragraph{HyperNetwork (tied case)}
In the tied case, we have one MLP that generates an output for each vector, so the number of operations needed for an MLP of input and hidden size $d$ and output sizes $d'$: $N(2 d^2 + 2dd' + 9d)$

\paragraph{Mixing MLP}
The mixing MLP has input and output size $N$ and hidden size $d'$, which is applied to each of the $d$ embedding dimensions (i.e., after transposition), incurring $d(4d'N+ 9')$ operations in total.

\paragraph{Total:}
The total number of operations in HyperMixer is
$d  (4 N d' + 9d') + N (2d^2 + 2 d'd + 9d)$

\subsection{Self-attention}
Multi-head self-attention with $h$ heads applies self-attention independently to each head consisting of vectors of size $d / h$, respectively.

Self-attention consists of
\begin{itemize}
    \item 3 linear layers to transform queries, keys, and values: $6h(d/h)^{2}$
    \item $h$ matrix multiplications with sizes $ N(d/h)$, totalling $2h(d / h)N^{2}$ operations
    \item softmax: $3N$
    \item a weighted average for each of the inputs, consisting of $(2dN^{2})$ operations.
\end{itemize}

In total: $6h(d/h)^{2}+ hN^{2}2(d / h) + 3N + (2dN^{2})$
\section{Connection with Lambda Layers and Linear Transformer}\label{sec:app:lambda}

We saw in Section~\ref{sec:results:locmix-attention} that HyperMixer was able to allow a form of attention without computing an attention matrix directly and thus scaling only linearly with the input length. In that regard, this method is similar to other methods such as \cite{bello2021lambdanetworks} or \cite{katharopoulos2020transformers}. We will describe here the difference between these approaches and our method.
Let us write the standard attention formula and the HyperMixer layer under the following form: 
\begin{equation}
    \text{Attention}(\boldsymbol{Q}, \boldsymbol{K}, \boldsymbol{V}) = \text{softmax}(\boldsymbol{QK}^{T})\boldsymbol{V}
\end{equation}

\begin{equation}
    \text{HyperMixer}(\boldsymbol{X}) = \boldsymbol{W}_1 \sigma(\boldsymbol{W}_2^T \boldsymbol{X})
\end{equation}
where $\boldsymbol{Q}, \boldsymbol{K}, \boldsymbol{V}, \boldsymbol{W}_1,\boldsymbol{W}_2 \in \mathbb{R}^{N \times d'}$, $\boldsymbol{X} \in \mathbb{R}^{N \times d}$ and  $\boldsymbol{W}_1,\boldsymbol{W}_2$ are the weights generated by the hypernetwork.

We can notice that the two operations differ mainly in the non-linearity location and the uses of linear or non-linear projection of the input.
Indeed, attention applies a non-linearity to $\boldsymbol{QK}^{T}$ and uses linear projection of the input $(\boldsymbol{Q}, \boldsymbol{K}, \boldsymbol{V})$ to construct the attention map.
On the contrary, HyperMixer uses two linear mapping of the input ($\boldsymbol{W}_1,\boldsymbol{W}_2$) and applies a non-linearity to $\boldsymbol{W}_2^T \boldsymbol{X}$, which is similar in a way to $\boldsymbol{K}^{T}\boldsymbol{V}$.
The quadratic cost of the attention layer comes from the place of the non-linearity as it requires the explicit computation of $\boldsymbol{QK}^{T} \in \mathbb{R}^{N \times N}$ which is quadratic with respect to the input size.
Most of the strategies used to overcome this quadratic cost generally find a way of moving this non-linearity.
This is the case of \cite{katharopoulos2020transformers} which applies non-linearities $\phi$ independently to $\boldsymbol{Q}$ and $\boldsymbol{K}$ and \cite{bello2021lambdanetworks} that applies softmax only to $\boldsymbol{K}$.
In that regard, these two methods can be compared with HyperMixer as they all scale linearly with the input size due to the non-linearity location.
Still, HyperMixer is conceptually different because it uses a non-linear transformation of the input and because it uses, in our opinion, a simpler and more understandable design entirely based on MLPs.

\section{Ablations on Transformer Layout}\label{appendix:transformer-layout}

While all Transformer layouts have a feature mixing and a token mixing component in each layer, the arrangement and connection of these components through skip connections and normalization layers remains an open question.
The original Transformer paper~\citep{vaswani2017attention} uses what is now known as the "post-norm" layout:
\begin{align*}
   & \mathbf{x}^1  &=&\; \operatorname{LayerNorm} (\mathbf{x} + \operatorname{token\_mixing}(\mathbf{x})) \\
   & \mathbf{x}^{out} &=&\; \operatorname{LayerNorm} (\mathbf{x^1} + \operatorname{feature\_mixing}(\mathbf{x^1}))
\end{align*}
where $\mathbf{x} \in \mathbb{R}^{N \times d}$ is the input to the layer, and $\mathbf{x}^{out} \in \mathbb{R}^{N \times d}$ is the output of the layer.

\cite{wang2019learning} proposes the "pre-norm" layout:
\begin{align*}
    &\mathbf{x}^1 &=&\; \mathbf{x} + 
 \operatorname{token\_mixing}(\operatorname{LayerNorm}(\mathbf{x})) \\
    &\mathbf{x}^{out} &=&\;  \mathbf{x^1} + \operatorname{feature\_mixing}(\operatorname{LayerNorm}(\mathbf{x^1}))
\end{align*}

\cite{bachlechner2021rezero} proposes the "ReZero" normalization, which introduces a learnable scalar $
\alpha \in \mathbb{R}$, initialized to zero:
\begin{align*}
    &\mathbf{x}^1 &=&\; \mathbf{x} + 
 \alpha_1\cdot\operatorname{token\_mixing}(\mathbf{x}) \\
    &\mathbf{x}^{out} &=&\;  \mathbf{x^1} + \alpha_2\cdot\operatorname{feature\_mixing}(\mathbf{x^1})
\end{align*}

\cite{wang2021gpt} observe that a speed-up can be obtained by parallelizing the two components:
\begin{align*}
    \mathbf{x}^{out} =  \mathbf{x} + &
 \operatorname{token\_mixing}(\operatorname{LayerNorm}(\mathbf{x})) \\
 + & \operatorname{feature\_mixing}(\operatorname{LayerNorm}(\mathbf{x}))
\end{align*}.

Finally, \cite{chowdhery2022palm} call the following the "standard serialized" formulation:
\begin{align*}
    &\mathbf{x}^1 & =&\; \mathbf{x} + 
 \operatorname{token\_mixing}(\operatorname{LayerNorm}(\mathbf{x})) \\
    &\mathbf{x}^{out} & =&\;  \mathbf{x} + \operatorname{feature\_mixing}(\operatorname{LayerNorm}(\mathbf{x^1})).
\end{align*}
As Figure~\ref{fig:model-architecture} shows, this is the model we have fixed for all previous experiments.

In the following, we combine each of the presented layouts with self-attention and HyperMixing, respectively.
Since we noticed early that the training with HyperMixing is not stable with some of the layouts, we also experimented with adding two different kinds of normalization to HyperMixer: layer normalization applied after $\tmmlp$, as shown in Algorithm~\ref{alg:hypermixer}, and length normalization.
For the latter, we simply scale the generated weight matrices by $\frac{1}{M}$, where $M$ is the number of keys.
The intuition is that this keeps the magnitude of activations in the hidden layer of $\tmmlp$ approximately the same across different input lengths.

\paragraph{Results}
Table~\ref{tab:results-skeleton} shows the best validation set results after tuning the learning rate using a logarithmically spaced grid of 7 values $\alpha \in \{0.001, 0.0005, 0.0002, 0.0001, 0.00005, 0.00002, \\ 0.00001\}$.

The results show that self-attention is relatively insensitive with respect to the type of layout, as all models except for ReZero attain an accuracy of 76-77\% on average.
In contrast, HyperMixer without normalization performs substantially worse with prenorm, ReZero, and the parallel layout. Length normalization mitigates this problem to some degree, but the addition of layer normalization yields the overall best results, where all models achieve between 77 and 78\% of accuracy on average.
We, therefore, recommend adding layer normalization by default when using HyperMixing in a new context.

\begin{table*}
    \centering
    \begin{tabular}{c|c|c|c|c|c|r}
    \hline
       \textbf{Layout} & \textbf{MNLI} & \textbf{SNLI} & \textbf{QQP} & \textbf{QNLI} & \textbf{SST} & \textbf{Average} \\
        \hline
             & \multicolumn{6}{c}{Multi-head self-attention}  \\
        \hline
serialized & 65.71 & 80.88 & 82.99 & 69.67 & 79.70 & 75.79\\
post-norm & 66.13 & 81.70 & 84.31 & 71.54 & 79.70 & 76.68\\
pre-norm & 66.60 & 80.59 & 82.96 & 73.13 & 80.73 & 76.80 \\
ReZero & 56.83 & 70.85 & 77.72 & 63.44 & 78.10 & 69.39\\
parallel & 66.30 & 81.46 & 83.12 & 71.55 & 79.70 & 76.43\\
    \hline
             & \multicolumn{6}{c}{HyperMixing (no normalization)}  \\
    \hline
serialized & 66.18 & 81.63 & 85.59 & 78.4 & 81.65 & 78.69\\
post-norm & 62.59 & 79.49 & 82.37 & 76.75 & 80.39 & 76.32\\
pre-norm & 56.62 & 78.49 & 82.88 & 64.18 & 81.08 & 72.65\\
ReZero & 35.45 & 33.82 & 63.18 & 49.46 & 49.08 & 46.20\\
parallel & 60.37 & 79.71 & 83.62 & 65.24 & 80.16 & 73.82\\
    \hline
             & \multicolumn{6}{c}{HyperMixing (length normalization)}  \\
    \hline
        serialized & 65.91 & 81.27 & 85.27 & 77.80 & 81.88 & 78.43\\
post-norm & 62.67 & 79.46 & 82.61 & 76.53 & 80.39 & 76.33\\
pre-norm & 64.83 & 80.71 & 84.41 & 76.31 & 81.65 & 77.58\\
ReZero & 35.45 & 33.82 & 63.18 & 70.31 & 54.13 & 51.38\\
parallel & 65.37 & 81.12 & 84.44 & 76.77 & 80.28 & 77.60 \\
    \hline
             & \multicolumn{6}{c}{HyperMixing (layer normalization)}  \\
    \hline
serialized & 66.47 & 81.36 & 85.74 & 77.72 & 80.50 & 78.36\\
post-norm & 64.26 & 80.05 & 83.81 & 76.62 & 80.85 & 77.12\\
pre-norm & 64.72 & 81.05 & 83.81 & 76.11 & 81.54 & 77.45\\
ReZero & 65.64 & 80.74 & 84.45 & 74.41 & 81.08 & 77.26\\
parallel & 65.49 & 80.59 & 84.43 & 76.53 & 81.65 & 77.74\\
    \end{tabular}
    \caption{Best validation set results on natural language understanding tasks after tuning the learning rate on a grid.}
    \label{tab:results-skeleton}
\end{table*}

\end{document}